\documentclass[10pt,twocolumn,letterpaper]{article}
\usepackage{cvpr}              
\usepackage{subcaption} 
\usepackage{graphicx} 
\usepackage{balance}
\definecolor{cvprblue}{rgb}{0.21,0.49,0.74}
\usepackage[pagebackref,breaklinks,colorlinks,allcolors=cvprblue]{hyperref}
\usepackage[table]{xcolor}
\usepackage{colortbl}
\newcommand{\chao}[1]{\textcolor{blue}{[Chao: #1]}}

\definecolor{qing}{RGB}{48, 192, 180}

\newcommand{\expert}[1]{\textcolor{myyellow}{v173}}
\newcommand{\plain}[1]{\textcolor{myblue}{5uus}}
\newcommand{\good}[1]{\textcolor{qing}{jvsZ}}

\usepackage{xcolor}
\definecolor{softpurple}{HTML}{8F7BFF}
\newcommand{\ClipGStream}{\textcolor{softpurple}{\textbf{ClipGStream}}}

\def\paperID{xxxx} 
\def\confName{xxxx}
\def\confYear{2026}

\def\ourname{ClipGStream}
\def\vrulong{Long 360}
\def\longNdv{flame salmon}

\def\staticInheritTrainingStrategy{Inter-clip Inheritance Strategy}
    \def\decoderInheritModule{Decoder Inheritance Module}
    \def\targetAnchorsInherit{Anchors Inheritance Module}
\def\dynamicIndepentTrainingStrategy{Intra-clip Training Strategy}
    \def\clipIndepentSpatioTemporalFields{Clip-Specific Spatio-Temporal Fields}
    \def\sourceAnchorsCompensationModules{Residual Anchors Compensation Module}
    
\def\firstClip{Reference Clip}
\def\otherClip{Source Clip}
\def\firstClipAbb{Reference Clip}
\def\otherClipAbb{Source Clip}

\def\framename{Clip-Stream}
\def\nonframename{Clip}
\def\streamframename{Frame-Stream}

\def\AnchorK{$A_0$}

\definecolor{myblue}{RGB}{72, 116, 203}
\definecolor{myyellow}{RGB}{238, 130, 47}
\definecolor{mycliporange}{HTML}{ED7D31}
\definecolor{myclipblue}{HTML}{4472C4}
\definecolor{myclipbrown}{RGB}{255, 192, 0}
\definecolor{myclipgreen}{RGB}{146, 208, 80}
\definecolor{streamblue}{RGB}{46, 84, 161}
\definecolor{clipbrown}{RGB}{255, 147, 0}

\title{\ClipGStream{}: Clip-Stream Gaussian Splatting for Any Length and Any Motion Multi-View Dynamic Scene Reconstruction} 

\author{
Jie Liang\textsuperscript{1, 2} *, Jiahao Wu\textsuperscript{1, 2} *, Chao Wang\textsuperscript{2} ‡, Jiayu Yang \textsuperscript{2}, Xiaoyun Zheng \textsuperscript{2}, \\ 
Kaiqiang Xiong \textsuperscript{1, 2}, Zhanke Wang \textsuperscript{1}, Jinbo Yan \textsuperscript{1}, FengGao \textsuperscript{3}, Ronggang Wang\textsuperscript{1, 2, 4} † 
\and 
\textsuperscript{1} Guangdong Provincial Key Laboratory of Ultra High Definition Immersive Media Technology,\\ 
Shenzhen Graduate School, Peking University 
\\ \textsuperscript{2} Pengcheng Laboratory, 
\textsuperscript{3} Peking University, 
\textsuperscript{4} MIGU Video Co., Ltd., \\
{\tt\small \{liangjie, wjh0616, xyun\_z, xiongkaiqiang, zk\_wang, yjb\}@stu.pku.edu.cn} \\
{\tt\small \{jiayuyang, gaof\}@pku.edu.cn\quad winchao1984@gmail.com\quad rgwang@pkusz.edu.cn}
}

\begin{document}

\twocolumn[{%
	\renewcommand\twocolumn[1][]{#1}%
	\maketitle
	\begin{center}
		\centering
        \vspace{-1.5em}
		\includegraphics[width=\textwidth, trim=0cm 11.5cm 10.3cm 0cm]{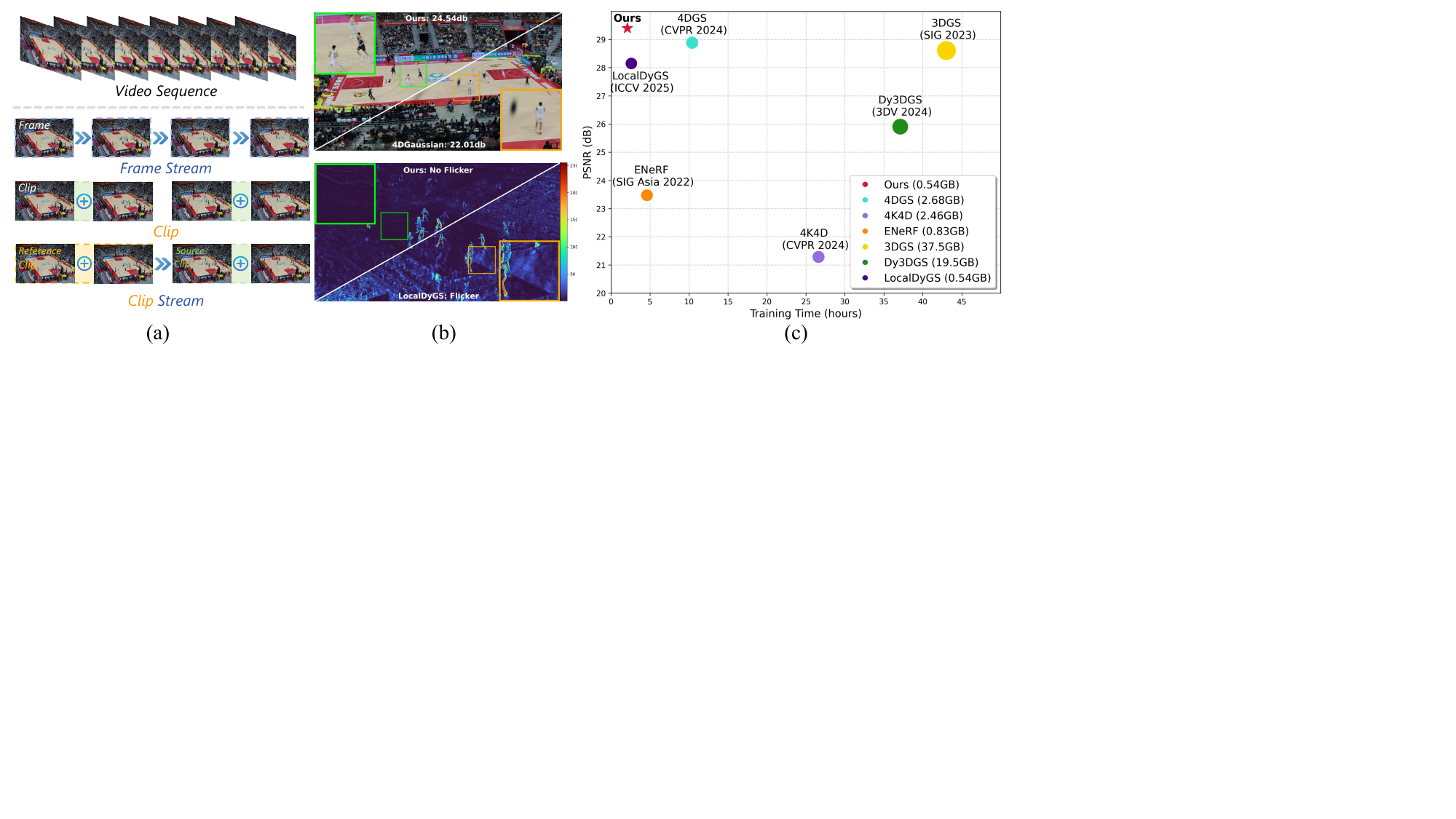}
		
		\captionof{figure}{\ourname~ enables scalable and temporally stable dynamic scene reconstruction for sequences of any length and any motion.
			(a) Unlike \textcolor{streamblue}{\streamframename~}methods that accumulate errors and \textcolor{clipbrown}{\nonframename~}methods that produce clip-level flicker, ours \textcolor{clipbrown}{Clip}-\textcolor{streamblue}{Stream} framework divides the video into a \textcolor{myclipbrown}{\firstClip} and subsequent \textcolor{myclipgreen}{\otherClip}s, where each \otherClip~is trained on top of the trained representation of the \firstClip~to ensure robust large-motion handling and temporal consistency.
			(b) \ourname~achieves higher reconstruction quality on the 1,400-frame \textit{\vrulong}~dataset. Cross-clip residual heatmaps show effective flicker suppression.
			(c) \ourname~further surpasses prior SOTA methods on the \textit{\longNdv} scene (1,200 frames) in both reconstruction fidelity and training efficiency.
		}
		\label{fig:teaserfig}
	\end{center}
}]

\renewcommand{\thefootnote}{\fnsymbol{footnote}}

\footnotetext[1]{These authors contributed equally to this work.}
\footnotetext[2]{Corresponding author.}
\footnotetext[3]{Co-corresponding author.}

\begin{abstract}

Dynamic 3D scene reconstruction is essential for immersive media such as VR, MR, and XR, yet remains challenging for long multi-view sequences with large-scale motion. Existing dynamic Gaussian approaches are either \streamframename, offering scalability but poor temporal stability, or \nonframename, achieving local consistency at the cost of high memory and limited sequence length.
We propose \mbox{\ourname}, a hybrid reconstruction framework that performs stream optimization at the clip level rather than the frame level. The sequence is divided into short clips, where dynamic motion is modeled using clip-independent spatio-temporal fields and residual anchor compensation to capture local variations efficiently, while inter-clip inherited anchors and decoders maintain structural consistency across clips. This \framename~design enables scalable, flicker-free reconstruction of long dynamic videos with high temporal coherence and reduced memory overhead. Extensive experiments demonstrate that \ourname~achieves state-of-the-art reconstruction quality and efficiency. The project page is available at: \textcolor{blue!50}{https://liangjie1999.github.io/ClipGStreamWeb/}.


\end{abstract}   
\section{Introduction}
\label{sec:introduction}

With the rapid advancement of immersive display technologies such as virtual reality (VR), mixed reality (MR), and extended reality (XR) \cite{rauschnabel2022xr, burdea2003virtual, speicher2019mixed}, users increasingly demand high fidelity and interactive three dimensional media experiences. Compared with static scenes \cite{barron2021mip,hu2023tri,xiong2025clmvsnetunsupervisedmultiviewstereo, barron2023zip,chen2022tensorf,peng2024structureconsistentgaussiansplatting,Deng_Wang_Wu_Liang_Ma_Hu_Wang_2026,barron2022mip}, producing dynamic immersive content is much more difficult in both data acquisition and reconstruction algorithms.

For data acquisition, synchronized multiview video capture provides an effective solution for recording real world dynamics, offering rich spatial and temporal information that forms a strong basis for high quality reconstruction \cite{jin2023capture,wang2022mixed,li2022neural,bansal20204d,lombardi2019neural,zitnick2004high}. 
However, dynamic scene reconstruction \cite{yan2025instant} remains highly challenging. Unlike static environments \cite{fridovich2022plenoxels,radl2024stopthepop,yu2024gaussian,ren2024octree} that mainly involve reconstructing geometry \cite{huang20242d,zhang2024rade, xiong2026intrinsicgeometryappearanceconsistencyoptimization} and appearance, dynamic scenes additionally require accurate modeling of temporal motion, which becomes difficult under large scale movements, complex interactions between humans and objects, and long sequence durations.

Current dynamic reconstruction methods can be broadly categorized into two groups: \streamframename~and \nonframename~approaches. \streamframename~methods optimize the scene on a per-frame basis, enabling scalable reconstruction of ultra-long dynamic sequences. Dynamic3DGS \cite{luiten2023dynamic} performs independent static reconstruction for each frame, and 3DGStream \cite{sun20243dgstream} models inter-frame motion through Neural Transformation Cache, yet such approaches often exhibit inter-frame jitter and accumulated reconstruction errors.
	
\nonframename~methods jointly optimize longer temporal clips (e.g., around 300 frames) using different spatio-temporal training strategy. 4DGS \cite{yang2023real}, 4DGaussian \cite{wu20244d}, and SpaceTimeGS \cite{li2024spacetime} effectively alleviate temporal inconsistency but incur substantial memory and computation costs, limiting scalability to long sequences. In addition, while both paradigms can handle subtle motion, they struggle to reconstruct dynamic scenes involving large or fast movements.

In this paper, we propose \ourname, a hybrid multiview dynamic reconstruction framework that integrates the advantages of \streamframename~and \nonframename~paradigms, while further enabling robust reconstruction of complex and large-scale motions through clip-level stream processing.
The full video is divided into $N$ short clips, each containing a continuous sequence of $M$ multiview frames. The first clip $Clip_0$, serves as the \firstClip, and the remaining clips $Clip_{n \in [1,,N-1]}$, are treated as \otherClip s.


Our approach comprises two complementary strategies that align with intra- and inter-clip levels.  
(1) \textbf{\dynamicIndepentTrainingStrategy:} 
The \firstClipAbb~decodes static and dynamic features of anchors into Temporal Gaussians ~\cite{Wu2025LocalDyGSMG}, followed by rasterization. 
For each \otherClipAbb, we augment the anchor set with residual anchors on top of the \firstClip's anchors to capture newly introduced or displaced structures, and train an independent Spatio-Temporal Field ($STF$) for that clip to model its local motion.  
(2) \textbf{\staticInheritTrainingStrategy:} 
To maintain temporal consistency across clips, each \otherClipAbb~inherits the anchors, static features, and decoder from the \firstClipAbb. 
All inherited components remain frozen during optimization, enabling stable reconstruction of arbitrarily long dynamic sequences.

In summary, our contributions are as follows:
	\begin{itemize}
		\item We propose the first \framename~dynamic reconstruction framework, which simultaneously addresses large-scale motion and long-sequence modeling in multi-view dynamic scenes.
    \item We design a two stage training strategy. Intra-clip training employs residual initialization with independent $STF$ for robust modeling of complex and fast motions, while Inter-clip inheritance freezes shared components to maintain temporal consistency and enable stable reconstruction of arbitrarily long video sequences.
	\item Extensive experiments on public benchmarks (N3DV, VRU) and newly released \vrulong~dataset demonstrate that our method achieves state-of-the-art performance. 

\end{itemize}

\section{Related Work}
\label{sec:related_work}

Dynamic scene reconstruction can be categorized into \textbf{monocular} and \textbf{multiview} settings. The monocular setting aims to reconstruct dynamic scenes from video sequences captured by a single moving camera \cite{yan2023nerf,wu2026pearpixelalignedexpressivehuman,yang2024deformable,park2021nerfies,attal2023hyperreel,song2023nerfplayer,wang2023mixed}, but its reconstruction quality is often limited by restricted viewpoints and insufficient geometric constraints. In contrast, the multiview setting leverages multiple synchronized cameras to capture rich spatial and temporal information, making it more suitable for producing high-quality dynamic content for commercial applications such as VR, AR, and XR. Therefore, this work focuses on multiview dynamic scene reconstruction, which has recently been actively explored in \cite{wuswift4d,li2022streaming,sun20243dgstream,xu20244k4d,xu2024grid4d,li2024spacetime}. Current dynamic scene reconstruction methods fall into two main paradigms: \textbf{\streamframename} and \textbf{\nonframename} methods.


\textbf{\streamframename~methods} train dynamic multiview videos in a frame-wise manner.  StreamRF \cite{li2022streaming} adopts Plenoxels \cite{yu2022plenoxels} for scene representation, enabling fast on-the-fly training. Dynamic3DGS \cite{luiten2023dynamic} extends this idea by optimizing simplified 3D Gaussian structures for high-quality novel view synthesis. 3DGStream \cite{sun20243dgstream} introduces a Neural Transformation Cache (NTC) to maintain temporal continuity between frames, while later methods such as iFVC \cite{tang2025compressing} and Hicom \cite{gao2024hicom} further enhance storage efficiency and reconstruction speed. These approaches are highly scalable and can process ultra-long dynamic sequences, yet they often suffer from inter-frame jitter and accumulated errors over time.


\begin{figure*}[ht]
\centering
	\includegraphics[width=\textwidth, trim=0 2.6cm 0mm 0]{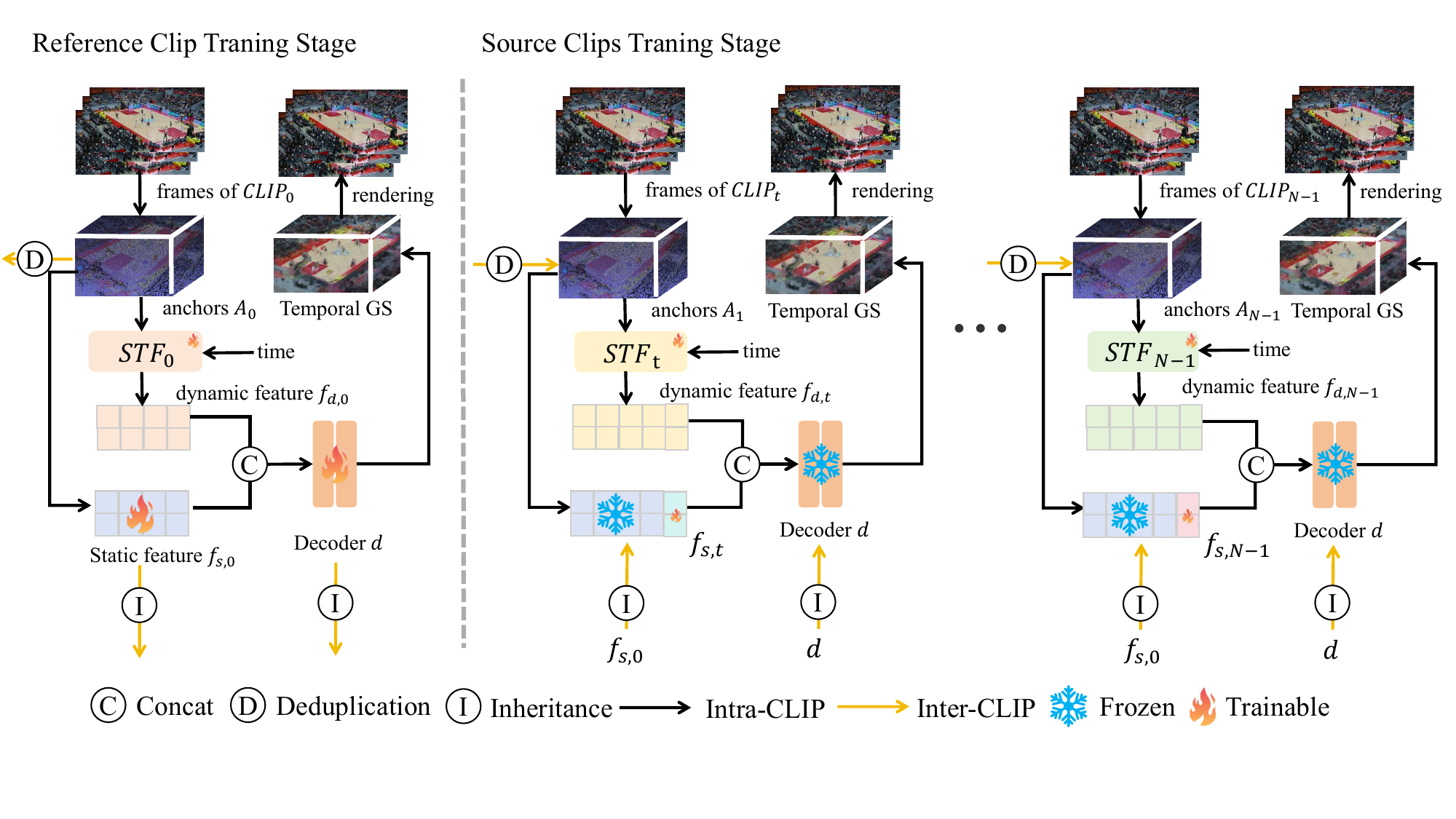}
  \caption{Our training process is divided into two stages: the \textbf{\firstClip}~($Clip_0$) Training Stage and \textbf{\otherClip}~($Clip_{n \in [1, N-1]}$) Training Stage. In \firstClip~Training Stage, anchors $A_0$ are initialized by the fused point cloud from all frames of $Clip_0$. The anchors are fed into a spatio-temporal field ($STF_0$) to extract dynamic features, which are then concatenated with static features. They are decoded to Temporal Gaussian attributes and rendered via rasterization. In \otherClip~Training Stage, we inherit the anchors, decoder and the static features from \firstClipAbb. Residual anchors are computed between the current and \firstClip~$A_0$. Finally, the same training pipeline as in the \firstClip~ stage is applied using the $A_n$ conpensated with residual anchors and the newly initialized $STF$.}

\label{fig:pipeline}
\vspace{-1.4em}
\end{figure*}


\textbf{\nonframename~methods} formulate dynamic scene reconstruction as a joint optimization problem over all frames, aiming to model coherent spatio-temporal structures within a unified representation.
K-Planes \cite{fridovich2023k} and HexPlanes \cite{cao2023hexplane} represent scene dynamics through learnable multi-plane features.
D-NeRF \cite{pumarola2021d}, Nerfies \cite{park2021nerfies}, 4D Gaussian \cite{wu20244d}, DeformableGS \cite{yang2024deformable}, Swift4D \cite{wuswift4d}, and SC-GS \cite{huang2024sc} model per-frame motion using continuous deformation fields to capture non-rigid dynamics in complex scenes \cite{kratimenos2023dynmf,lin2024gaussian}.
4DGS \cite{yang2023real} further adopts an explicit 4D representation of geometry and appearance, while SpaceTimeGS \cite{li2024spacetime}, Gaussian-Flow \cite{gao2024gaussianflow}, and NSFF \cite{li2020neural} estimate continuous motion trajectories to model long-range temporal dynamics.
Although these methods achieve temporally stable and high-quality reconstructions, their joint optimization over all frames prevents them from scaling to long dynamic sequences.
As a result, training is typically performed on short clips, and extending them to longer sequences often leads to temporal discontinuities across clip boundaries.

To enable high-quality reconstruction of long-sequence dynamic scenes with complex motions, we propose a novel \textbf{\framename} framework, \textbf{\ourname}.
Within each clip, a clip training strategy is employed to maintain temporal consistency and ensure stable reconstruction, while across clips, a stream-based scheme enables continuous and scalable learning.
By integrating the advantages of both paradigms, \ourname~unifies clip and stream modeling within a single coherent framework, achieving robust performance on ultra-long sequences and large-scale dynamic motions.


\section{Method}
\label{sec:method}

%
In this section, we present our method, \ourname.
We first review the preliminary concepts of 3DGS and ScaffoldGS in \cref{method:preliminary}. In \cref{method:framework}, we describe how our \mbox{\framename} framework combines the strengths of both \streamframename~and \nonframename~approaches, achieving scalable and temporally consistent dynamic scene reconstruction.
Next, Sec.~\ref{method:dynamic} introduces the \textbf{\dynamicIndepentTrainingStrategy}, in which the \firstClip~and each \otherClip~are modeled using independent Spatio-Temporal Fields ($STF$), and large motions in \otherClip~are compensated via residual anchors.
Following that, \cref{method:static} presents the \textbf{\staticInheritTrainingStrategy}, which shares the decoder and features of anchors across Clips to enforce temporal consistency. Finally, the overall loss is provided in \cref{method:loss}.



\subsection{3DGS and ScaffoldGS Preliminary}
\label{method:preliminary}

3D Gaussian Splatting (3DGS) \cite{kerbl20233d} has become a prominent technique for novel view synthesis by employing 3D Gaussians as fundamental rendering primitives. Each primitive is represented as \( G\{\mu, q, s, \sigma, c\} \), where the parameters represent mean (\(\mu\)), rotation (\(q\)), scaling (\(s\)), opacity (\(\sigma\)), and color (\(c\)), respectively. The 3D Gaussian function \( G(x) \) is formally defined as:
\begin{equation} 
	G(x) = e^{-\frac{1}{2}(x-\mu)^T\Sigma^{-1}(x-\mu)},
\end{equation} 

\begin{figure}[t]
	\centering

    \includegraphics[width=0.6\textwidth, trim=1cm 13cm 11cm 1cm]{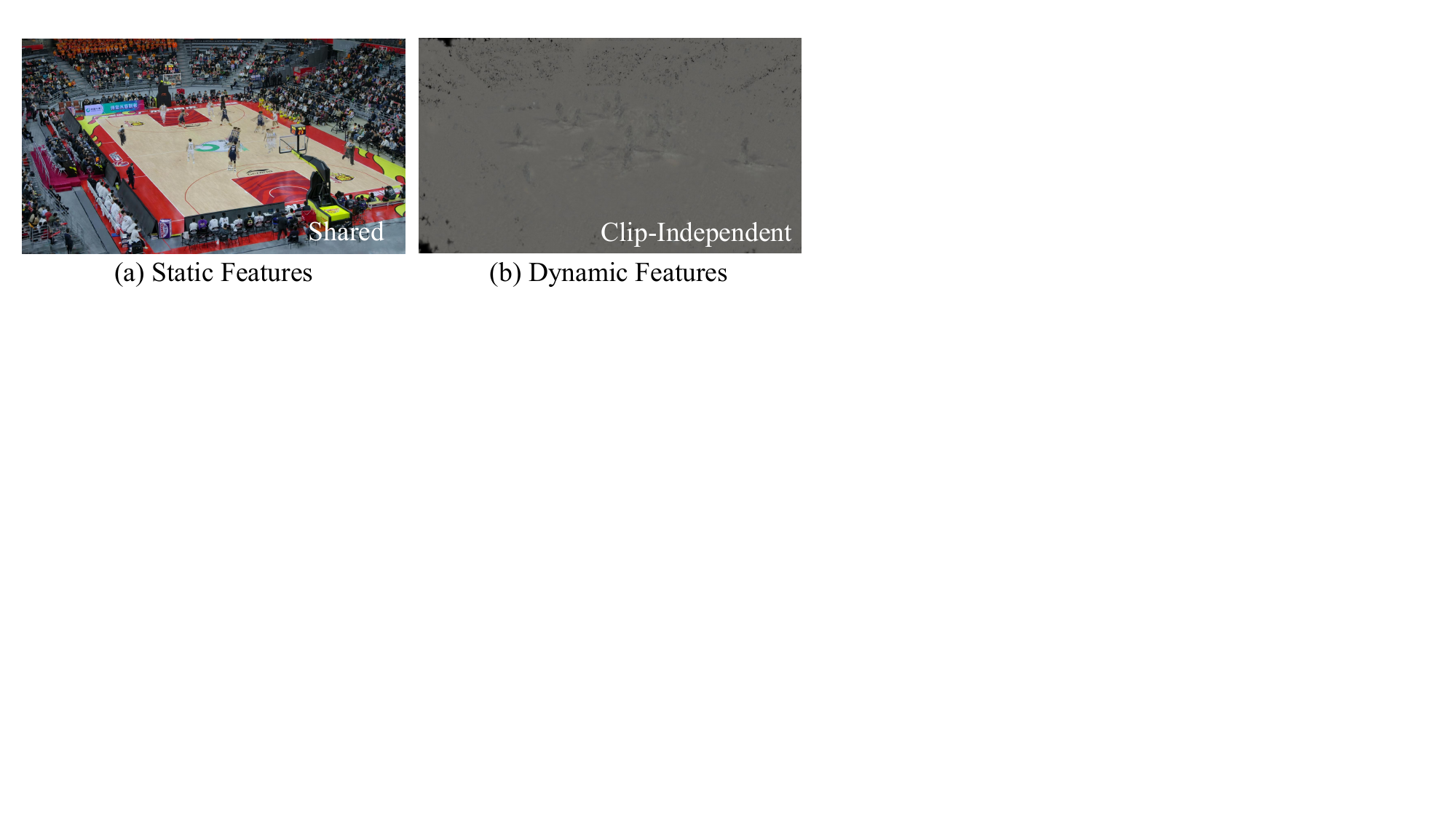}

	
    
	\caption{(a)Static features $f_s$ learns all background information, therefore sharing $f_s$ across clips ensures temporal consistency. (b) Dynamic features learns residual information that controls the visibility of dynamic content, and thus are clip-independent.}
	\vspace{-1.5em}
	\label{fig:feat}
\end{figure}

where \(\Sigma\) denotes the covariance matrix of the 3D Gaussian, which is usually computed from the rotation \(q\) and scaling \(s\).  In the rendering process, following the method in \cite{zwicker2002ewa}, each 3D Gaussian is projected onto the image plane as a 2D Gaussian \( G'(x) \). The rasterizer then sorts these 2D Gaussians and performs \(\alpha\)-blending to determine pixel colors.
\begin{equation}
	\resizebox{.9\linewidth}{!}{$
		C(p)=\sum_{i\in K}c_i\alpha_i(p)\prod_{j=1}^{i-1}(1-\alpha_j(p)),~~\alpha_i(p)=\sigma_iG_i'(p).
		$}
\end{equation} 
In this equation, \( p \) indicates the pixel location, and \( K \) represents the total number of 2D Gaussians that cover the given pixel. This differentiable rendering framework allows for end-to-end training using supervised multi-view images.


Although 3DGS achieves high rendering quality, its independent optimization of Gaussians leads to redundant primitives and limited representation compactness. ScaffoldGS \cite{lu2024scaffold} addresses this by organizing Gaussians around a set of learnable anchors. Each anchor is characterized by four attributes: a mean position $\mu_a \in \mathbb{R}^{3}$, a feature vector $f_a \in \mathbb{R}^{32}$, a scale factor $l_a \in \mathbb{R}^{3}$, and offsets $O_a \in \mathbb{R}^{k \times 3}$ associated with $k$ neural Gaussian primitives. The positions of the neural Gaussians are then computed as
\begin{equation}
    \{\mu_0,...,\mu_{k-1}\}=\mu_a+\{\mathcal{O}_0,...\mathcal{O}_{k-1}\}\cdot l_a.
\end{equation}
The neural Gaussian parameters corresponding to each anchor are decoded through MLPs. Our method also adopts a similar strategy of decoding features associated with anchors to obtain Gaussian parameters.

\subsection{\ourname}
\label{method:framework}
To address the limitations of Frame-Stream and Clip paradigms, we propose a unified \textbf{\framename}~framework, \ourname, that divides a dynamic sequence into temporally coherent clips.
The sequence begins with a \textbf{\firstClip}, which is fully optimized to establish a stable spatio-temporal representation, followed by multiple \textbf{\otherClip s} that are trained based on it.
This design combines the stability of clip-based optimization with the scalability of stream-based training, forming the foundation for our intra-clip and inter-clip strategies introduced below.

\subsubsection{\dynamicIndepentTrainingStrategy}
\label{method:dynamic}

In the \firstClip~training stage, as shown in \cref{fig:pipeline}, we formulate dynamic scene modeling as decoding both static and dynamic features for each anchor~\cite{Wu2025LocalDyGSMG}. The decoupling of static and dynamic features enables us to inherit the static information in subsequent Source Clips training, thereby ensuring temporal consistency (\cref{fig:feat}).
\begin{figure}[t]
	\centering
	\subfloat[Point → Field]{
		\includegraphics[width=0.23 \textwidth]{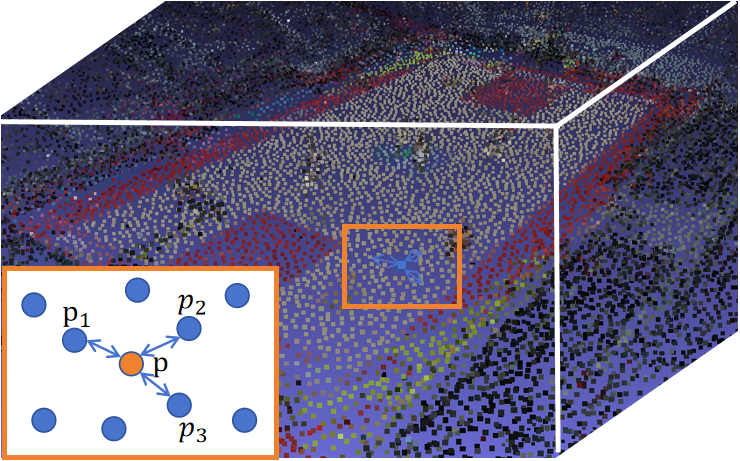}}
	\subfloat[Field → Residual]{
		\includegraphics[width=0.24 \textwidth]{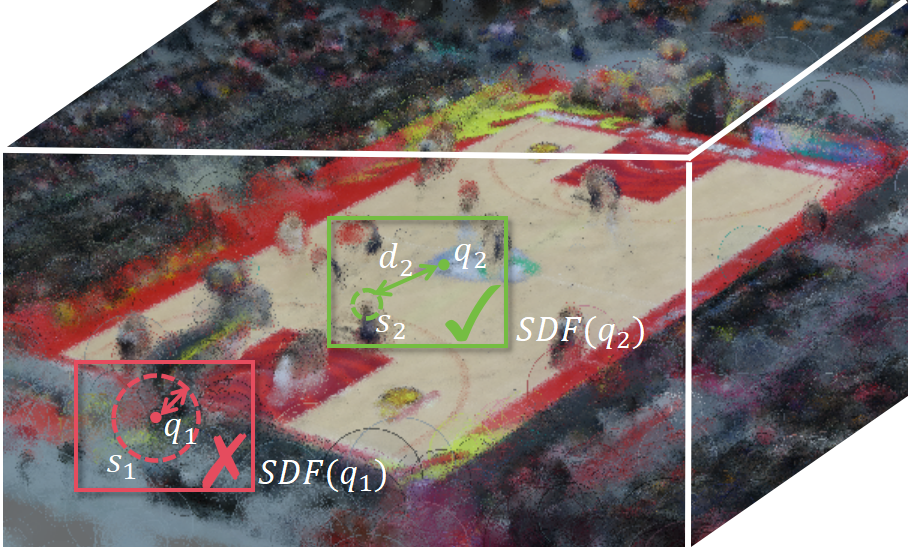}}
	\caption{\textbf{Geometry-aware deduplication.} (a)~The points from the \firstClipAbb~is converted into a \textit{spherical coverage field}: for each point $ p $, a sphere $s$ is centered at $ p $ with radius set to the mean distance to its three nearest neighbors (computed via KNN). 
		(b)~Given the field and the \otherClipAbb~candidate anchors, residual anchors are selected based on their signed distance to the field surface. $ q_1 $ (red box), $SDF(q_1) < 0$, is filtered out; while $ q_2 $ (green box), $SDF(q_2) > 0$, is retained as a residual anchor.} 
	\label{fig:dynamicPoint}
	\vspace{-1em}
\end{figure}

Specifically, the anchor set \AnchorK~is initialized from the COLMAP~\cite{schonberger2016structure, highfidelity} point cloud aggregated over all frames within $Clip_0$. 
Unlike ScaffoldGS~\cite{lu2024scaffold}, our anchors contain only a 3D position $\mu \in \mathbb{R}^{3}$ and two feature vectors: a static feature $f_s \in \mathbb{R}^{64}$ and a dynamic feature $f_d \in \mathbb{R}^{64}$, where the latter is obtained from a spatio-temporal field ($STF$). 
The $STF_0$ is implemented as a 4D hash grid $h_0$ followed by a fully fused MLP $\phi_0$. 
Given an anchor $\mu_0$ at time $t$, the dynamic feature is computed as
\begin{equation}
    f_{d,0} = \phi_0(h_{0}(\mu_{0}, t)).
\end{equation}
We then concatenate $f_{d,0}$ with its static counterpart $f_{s,0}$ and feed them into the decoder $d(\cdot)$ to generate the Temporal Gaussians
\begin{equation}
    G_{t,0} = d([f_{s,0}; f_{d,0}]),
\end{equation}
which are subsequently rasterized and supervised by ground-truth images. 
This process jointly optimizes the anchor positions $\mu_0$, the static features $f_{s,0}$, and the parameters of $STF_0$.
However, directly extending this process to subsequent \otherClipAbb~is challenging in dynamic scenes with large-scale motion. To maintain temporal consistency across clips, we first inherit $A_0$ when training the \otherClip~(as detailed in \cref{method:static}).
Anchors inherited from the \firstClipAbb~often undergo substantial spatial displacements that cannot be accurately captured through deformation alone. 
To handle such cases, we introduce a residual-based \textbf{anchor compensation strategy}, which augments the input anchor set with newly emerging or displaced anchors, thereby improving spatial coverage and motion adaptability. 
Specifically, we construct the anchor set $A_n$ for \otherClip~based on the residual between the COLMAP-derived anchors $A_n^c$ of $Clip_n$ and the base anchors $A_0$ from \firstClip:
\begin{equation}
    A_n = A_0 \cup A_n^r = A_0 \cup \operatorname{Dedup}(A_n^c, A_0),
\end{equation}
where $\operatorname{Dedup}(A_n^c, A_0)$ removes duplicate anchors that already exist in $A_0$. 
To achieve geometry-aware deduplication, each anchor $p \in A_0$ is represented by a sphere centered at $p$ with radius $r$ defined as the mean Euclidean distance to its three nearest neighbors $(p_1, p_2, p_3)$:
\begin{equation}
	r = \tfrac{1}{3} \sum_{i=1}^{3} \| p_i - p \|_2.
\end{equation}

As shown in \cref{fig:dynamicPoint}, this forms a \textit{spherical coverage field} describing the region already represented by $A_0$. 
For each candidate $q \in A_n^c$, its signed distance \cite{Zhou2018} to the coverage surface is computed using Signed Distance Fuction $SDF(\cdot)$; if $SDF(q) > 0$, $q$ is retained as a residual anchor in $A_n^r$, otherwise discarded. 
This procedure preserves newly introduced or significantly displaced structures while preventing redundant anchor growth.

Moreover, the dynamic features $f_{d,0}$ of anchors vary substantially across clips, and sharing a single $STF_0$ leads to inconsistent motion representations. 
To mitigate this problem, we allocate an \textbf{independent spatio-temporal field} to each clip, such as $STF_1, \ldots, STF_{N-1}$. 
This clip-specific design enables localized motion modeling while maintaining temporal coherence across clips, and significantly improves the reconstruction quality in long sequences, as validated in \cref{table:abl_smallGops}.

\subsubsection{\staticInheritTrainingStrategy}
\label{method:static}

For the \firstClipAbb, the training jointly optimizes the anchors $A_0$, their static features $f_{s,0}$, the spatio-temporal field $STF_0$, and the decoder $d$ that maps feature representations to Temporal Gaussian attributes. 

\begin{figure}[t]
	\centering
	\setcounter{subfigure}{0} 
		\subfloat[w/o RAC]{	
		\hspace*{0.001em}
        \includegraphics[width=0.48\linewidth]{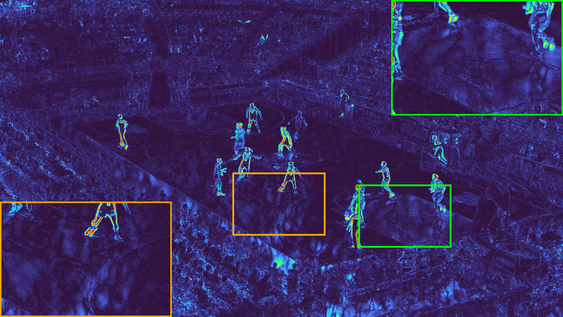}}
		\subfloat[w/o AI]{ 
		\includegraphics[width=0.515\linewidth]{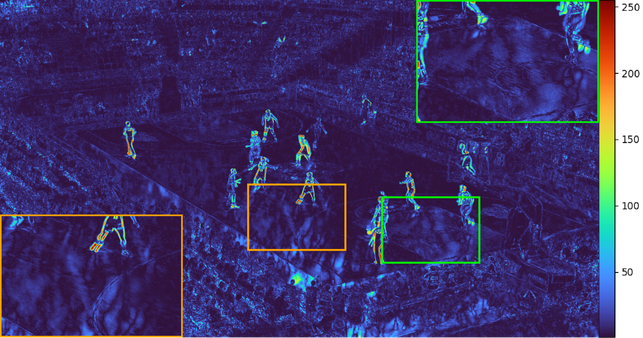}}
	\hfill
		\subfloat[w/o both]{
		\includegraphics[width=0.48\linewidth]{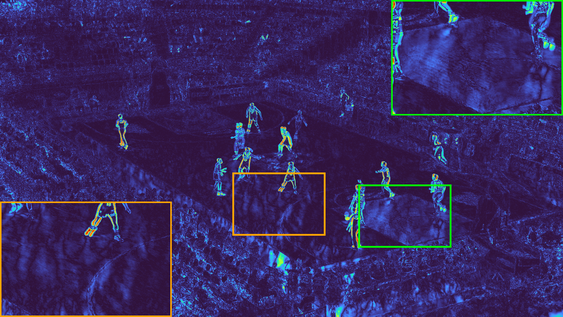}} 
		\subfloat[\textbf{ours}]{
		\includegraphics[width=0.48\linewidth]
        {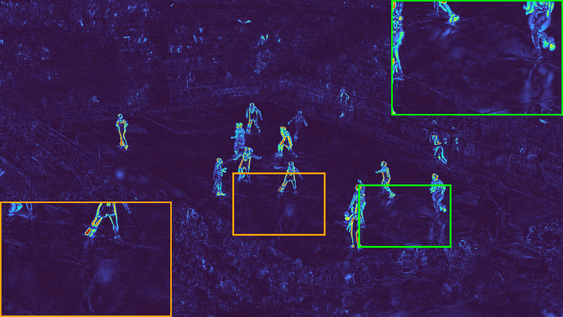}}

     \caption{\textbf{The ablation study on the \sourceAnchorsCompensationModules~(RAC) and the \targetAnchorsInherit~(AI)}. As seen from the residual heatmaps between adjacent clips, removing either module leads to strong responses in static regions as shown in (a)(b)(c), while enabling them, as illustrated in (d), significantly suppresses flicker and preserves smooth clip transitions, which demonstrates that both components play essential roles in maintaining inter clip stability.}
	\label{fig:errorsMaps}
	\vspace{-1em}
\end{figure}

However, naively independently optimizing each subsequent \otherClipAbb~($Clips_{n \in [1, N-1]}$) would require reinitializing these components, leading to inconsistent scene representations across clips. 
As shown in \cref{fig:errorsMaps}, such independent optimization results in severe flickering in static regions and degraded rendering quality in dynamic areas (\cref{fig:abl_optimizer}~(a)). 
To prevent this, we propose a \textbf{static inheritance strategy} that preserves the stable representation learned from the \firstClipAbb~and ensures temporal consistency across clips through anchor and decoder inheritance.

In this strategy, the anchors $A_0$ and their static features $f_{s,0}$ are inherited from the \firstClipAbb~and kept fixed during the training of all subsequent clips. 

This shared anchor foundation serves as a stable reference for static structures, preventing local re-optimization and ensuring temporal alignment throughout the sequence, as illustrated in \cref{fig:feat}~.
Meanwhile, the decoder $d$ trained on the \firstClipAbb~is reused by all \otherClipAbb~and remains frozen during their optimization, which guarantees a consistent decoding of geometry and appearance attributes across clips, significantly improving reconstruction quality (\cref{fig:abl_optimizer}(b), \cref{table:abl_optimizer}).

Building upon these inherited components, the training of each \otherClip~proceeds in a consistent manner. 
For $Clip_1$, we first apply the \sourceAnchorsCompensationModules~to generate residual anchors $A_1^r$ by deduplicating the COLMAP-derived anchors $A_1^c$ of $Clip_1$ against $A_0$, capturing newly introduced or displaced structures caused by large motion. 
Following the same residual compensation process described in Eq. (6), the anchor set $A_1$ is then obtained by merging the inherited anchors $A_0$ with the deduplicated anchors from $A_1^c$:
\begin{equation}
    A_1 = A_0 \cup \operatorname{Dedup}(A_1^c, A_0),
\end{equation}
where $\operatorname{Dedup}(A_1^c, A_0)$ removes anchors already covered by $A_0$. The anchors $A_0$ and their static features $f_{s,0}$ remain frozen, providing a stable reference for static regions, while the static feature of the new clip incorporates a learnable residual component $f_{s,1}^r$ associated with $A_1^r$:
\begin{equation}
    f_{s,1} = [f_{s,0}; f_{s,1}^r]
\end{equation}

For each anchor $\mu_1 \in A_1$ and timestamp $t$, the clip-specific spatio-temporal field $STF_1$ produces dynamic features as
\begin{equation}
    f_{d,1} = \phi_1(h_{1}(\mu_1, t)).
\end{equation}
The concatenated feature $[f_{s,1}; f_{d,1}]$ is then fed into the frozen decoder $d(\cdot)$ to generate Temporal Gaussians:
\begin{equation}
    G_{t,1} = d([f_{s,1}; f_{d,1}]).
\end{equation}

After rasterization, the rendered results are supervised by ground-truth images using photometric and regularization losses, while all inherited components ($A_0$, $f_{s,0}$, and $d$) remain frozen to ensure inter-clip consistency throughout the sequence.

\subsection{Loss Function}
\label{method:loss}


To promote compact Temporal Gaussians and constrain each Gaussian to represent only its local spatial region, we adopt a lightweight volume regularization term \(L_v\) \cite{lombardi2021mixture,lu2024scaffold}:
\begin{equation}
    L_v = \sum_{i=1}^{M} \text{Prod}(s_t^i),
\end{equation}
where \(M\) is the number of active Temporal Gaussians and \(s_t^i\) denotes the scale of the \(i\)-th Gaussian at time \(t\). 
Following the 3DGS formulation, we further include \(L_1\) and \(L_{SSIM}\) losses to enhance reconstruction quality. 
The overall training objective is given by
\begin{equation}
    L = (1 - \lambda_{SSIM})L_1 + \lambda_{SSIM}L_{SSIM} + \lambda_v L_v.
\end{equation}

\begin{figure*}[t]
    \centering

    \begin{minipage}[b]{0.25\textwidth}
        \centering
        \includegraphics[width=0.98\linewidth]{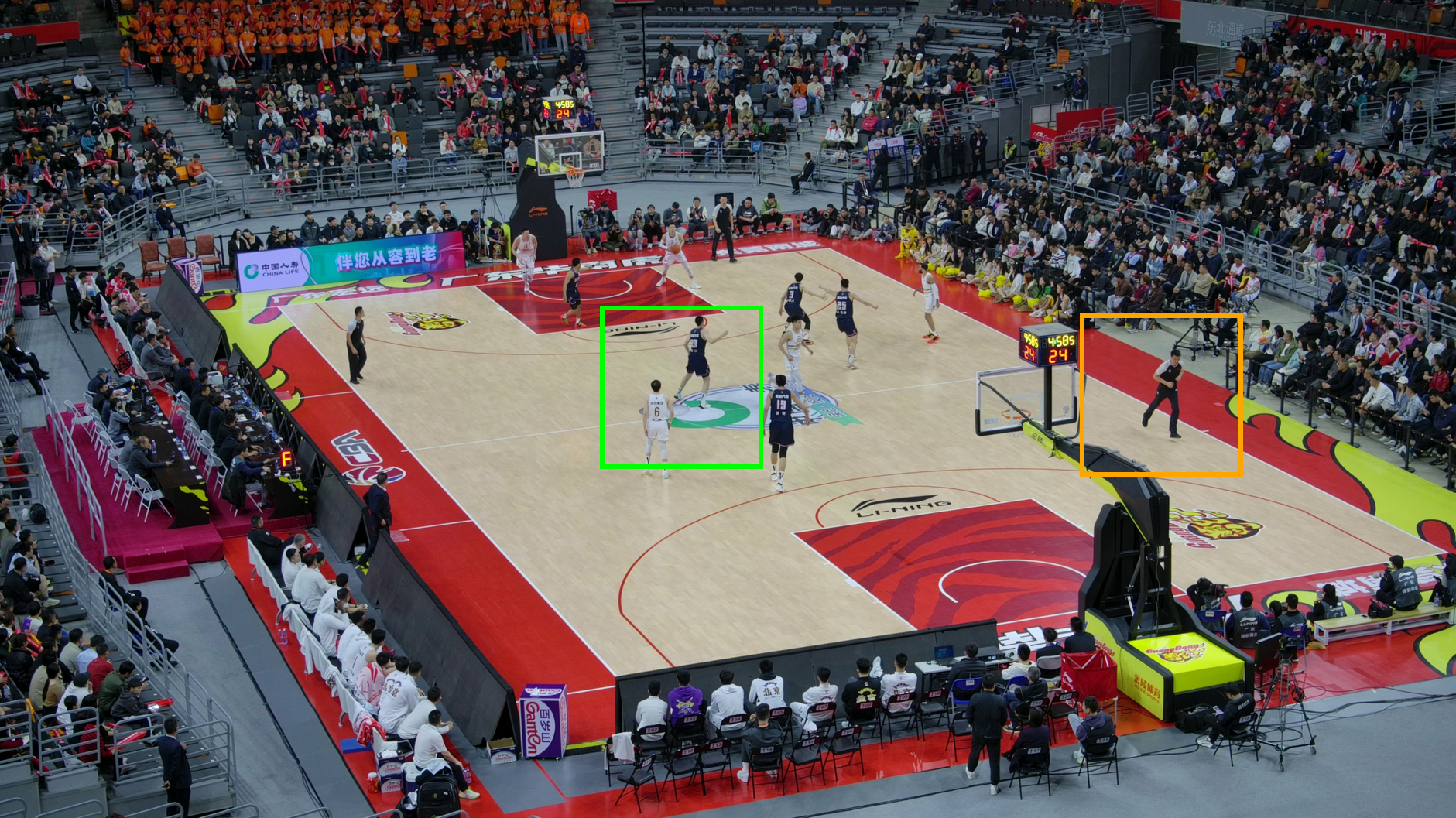} \\
        \includegraphics[width=0.475\linewidth]{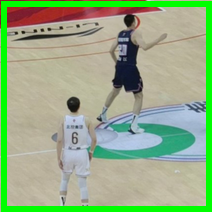}
        \includegraphics[width=0.475\linewidth]{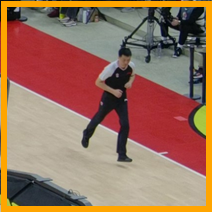} \\
        \includegraphics[width=0.98\linewidth]{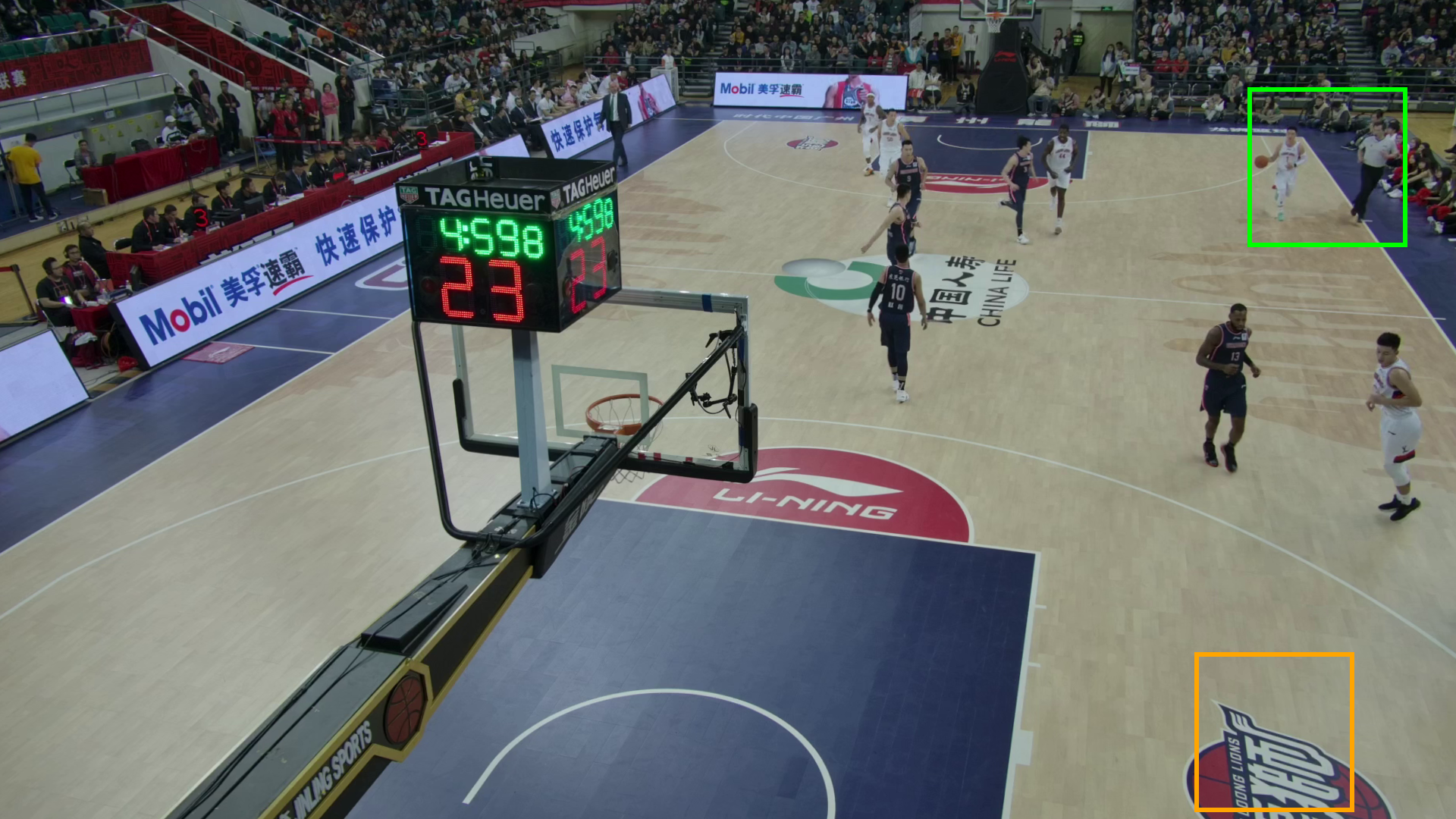} \\
        \includegraphics[width=0.475\linewidth]{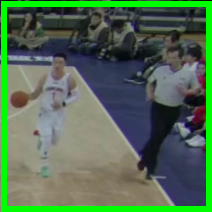}
        \includegraphics[width=0.475\linewidth]{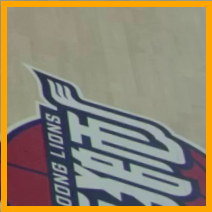}
        \caption*{(a) GT}
    \end{minipage}%
    \hfill
    \begin{minipage}[b]{0.25\textwidth}
        \centering
        \includegraphics[width=0.98\linewidth]{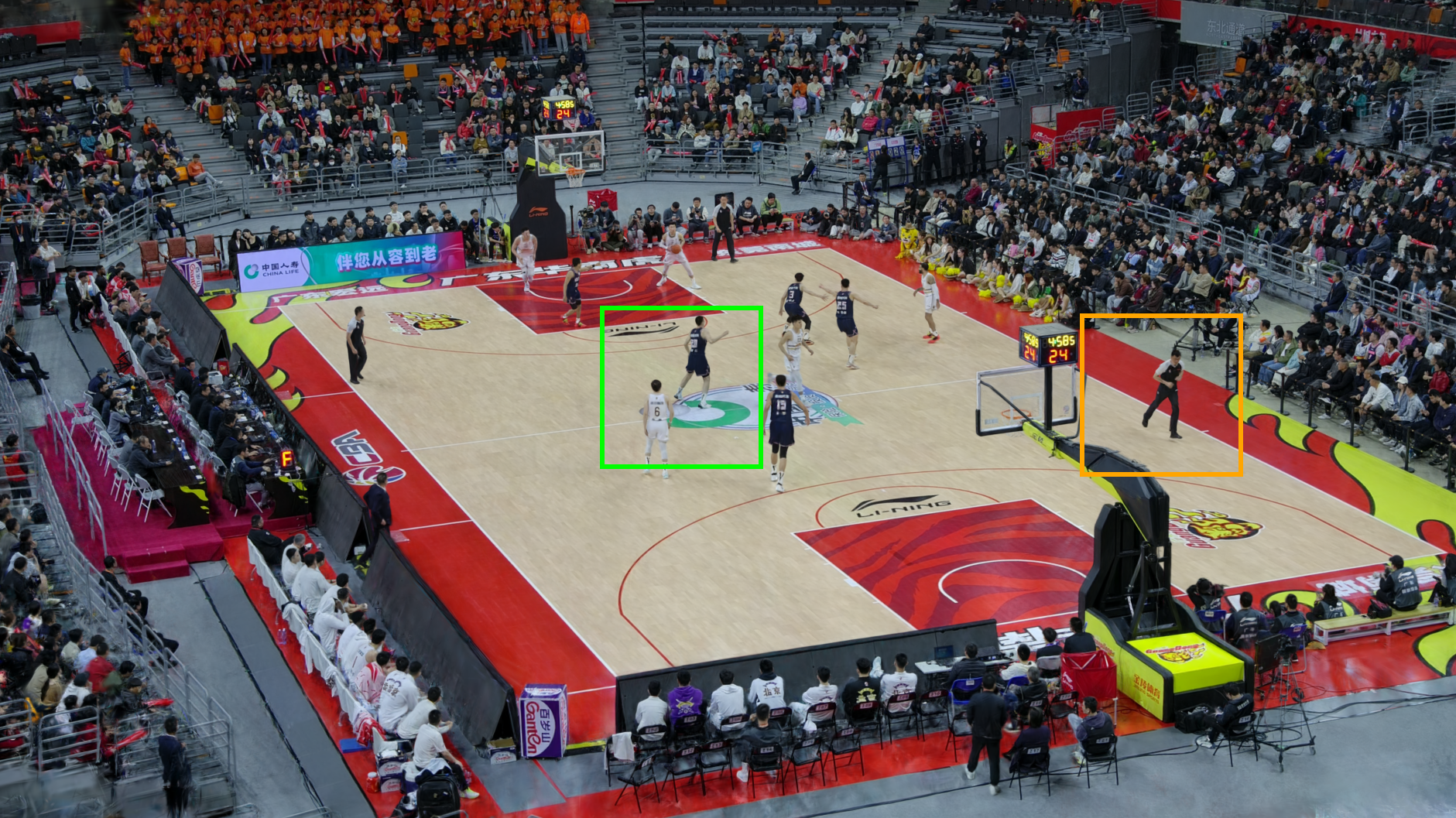} \\
        \includegraphics[width=0.475\linewidth]{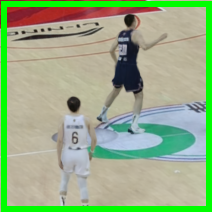}
        \includegraphics[width=0.475\linewidth]{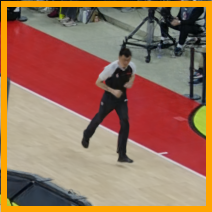} \\
        \includegraphics[width=0.98\linewidth]{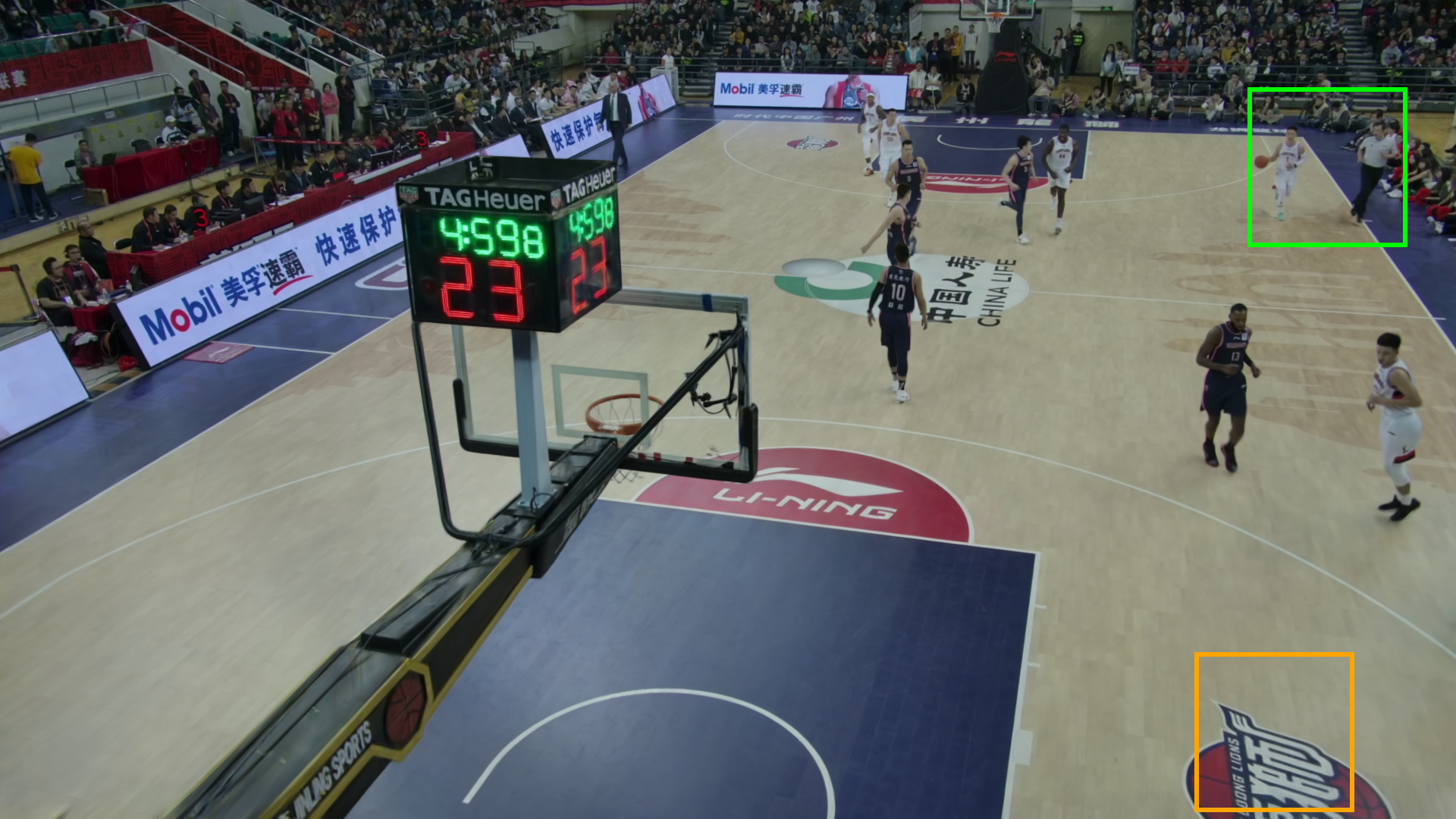} \\
        \includegraphics[width=0.475\linewidth]{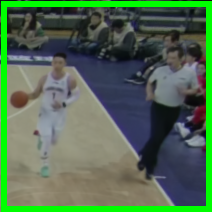}
        \includegraphics[width=0.475\linewidth]{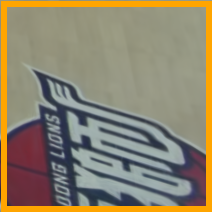}
        \caption*{(b) \textbf{Ours}}
    \end{minipage}%
    \hfill
    \begin{minipage}[b]{0.25\textwidth}
        \centering
        \includegraphics[width=0.98\linewidth]{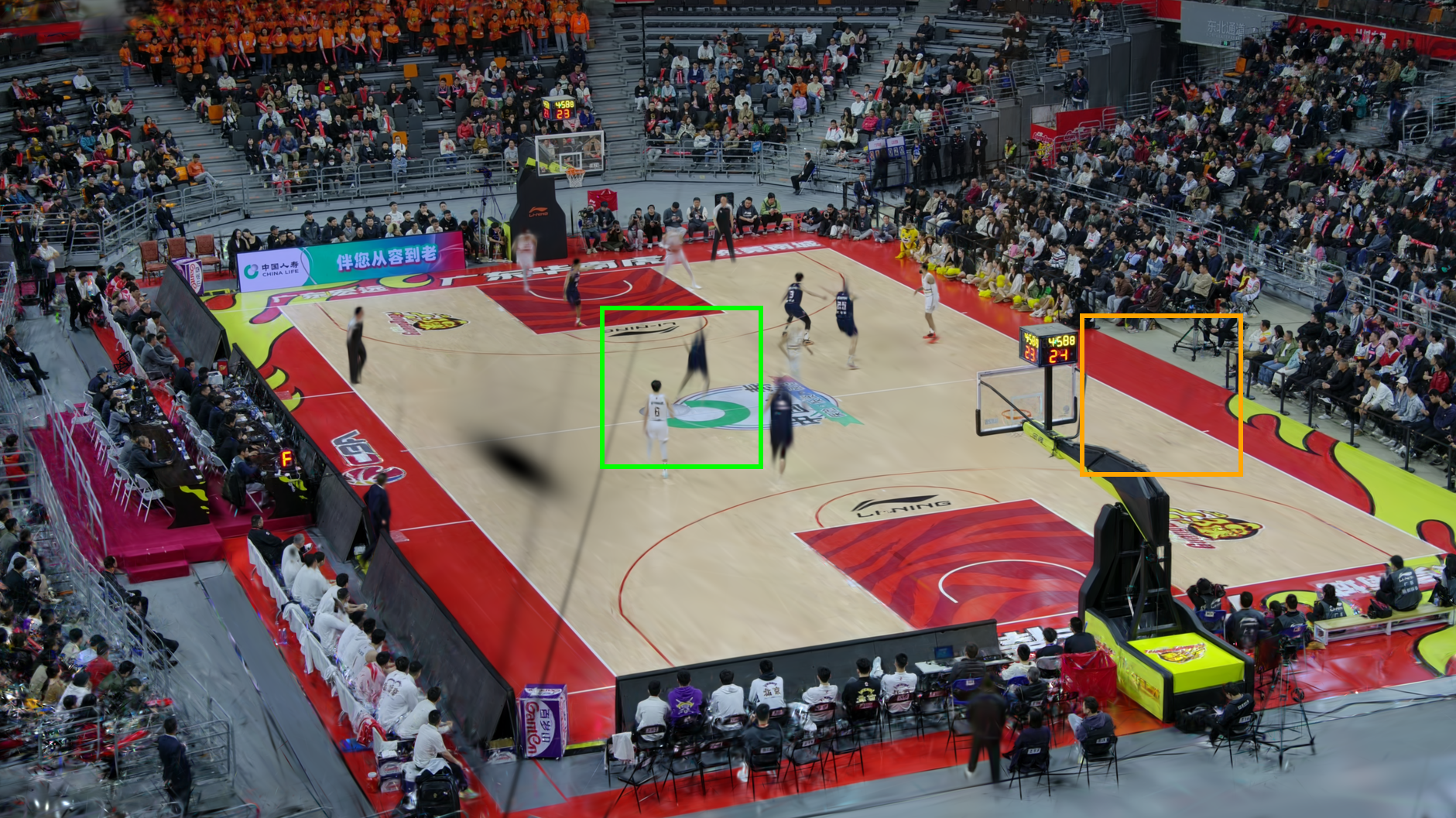} \\
        \includegraphics[width=0.475\linewidth]{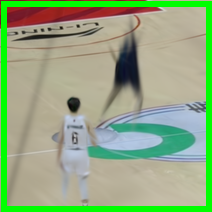}
        \includegraphics[width=0.475\linewidth]{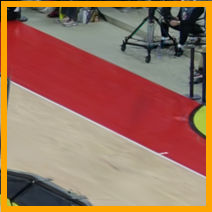} \\
        \includegraphics[width=0.98\linewidth]{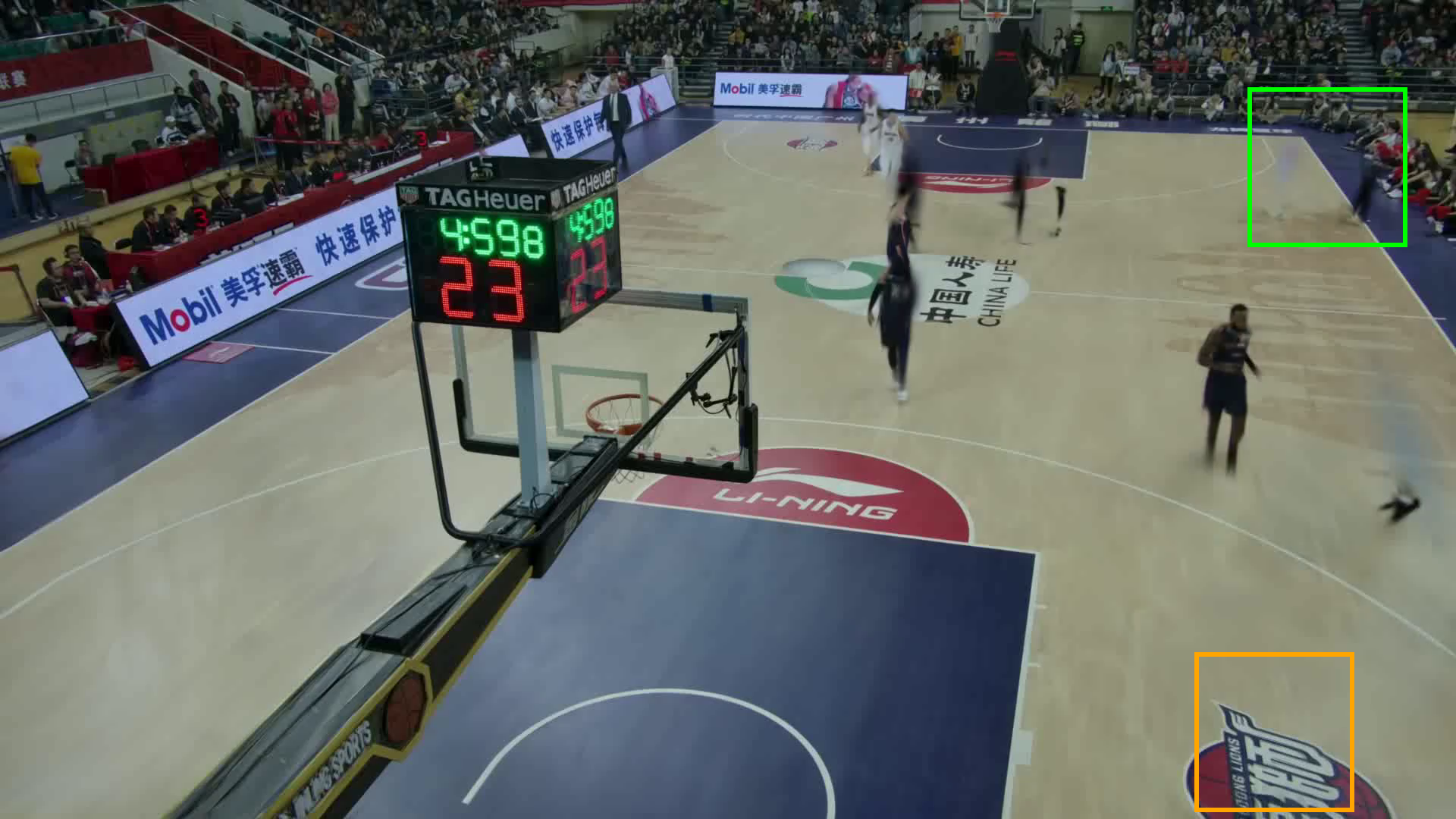} \\
        \includegraphics[width=0.475\linewidth]{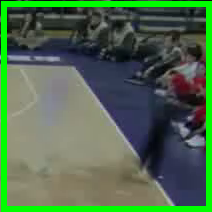}
        \includegraphics[width=0.475\linewidth]{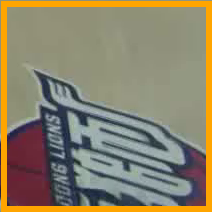}
        \caption*{(c) 4DGaussain \cite{wu20244d}}
    \end{minipage}%
    \hfill
    \begin{minipage}[b]{0.25\textwidth}
        \centering
        \includegraphics[width=0.98\linewidth]{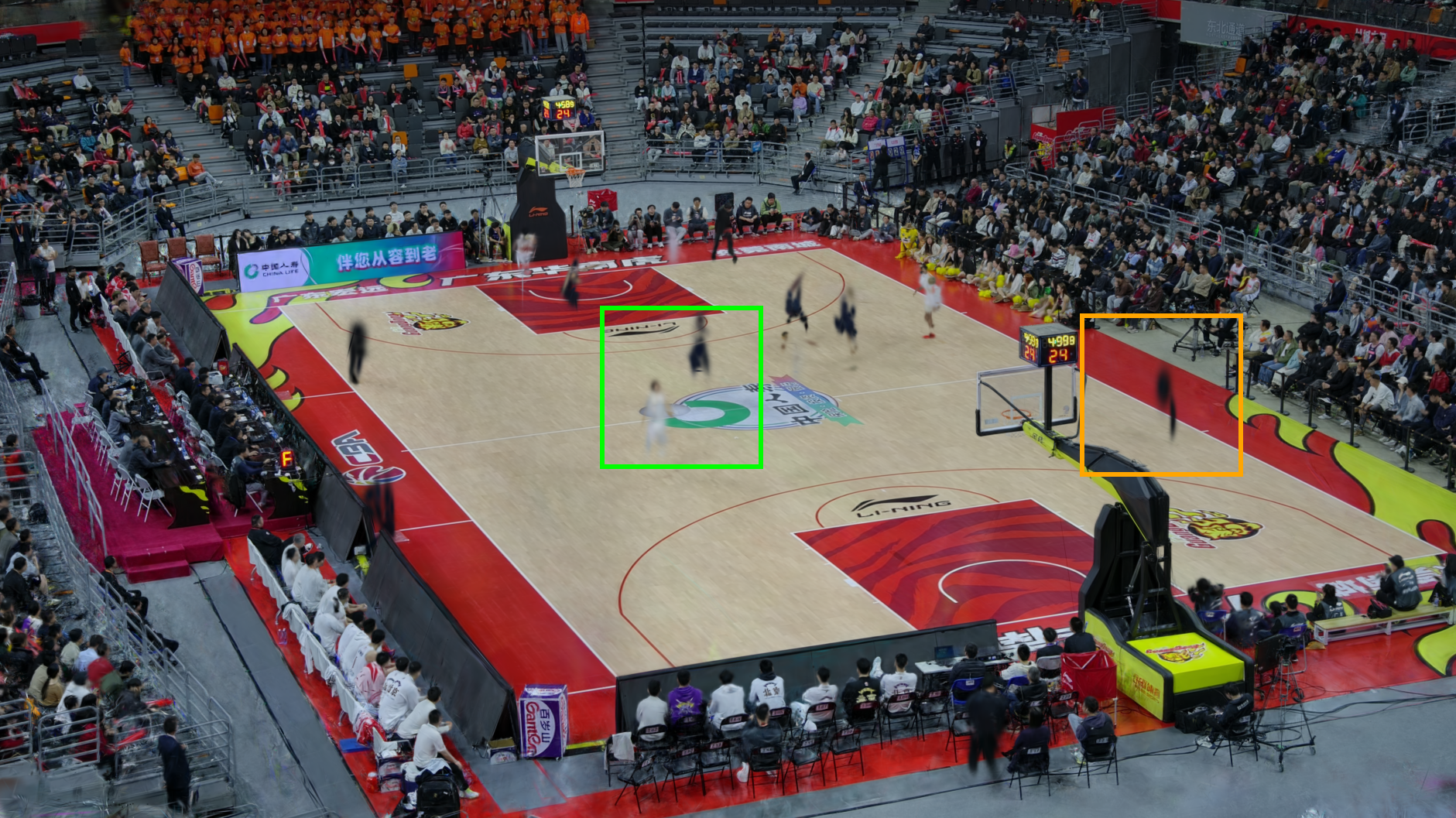}\\
        \includegraphics[width=0.475\linewidth]{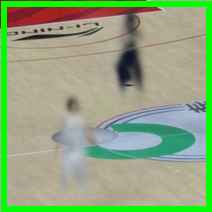}
        \includegraphics[width=0.475\linewidth]{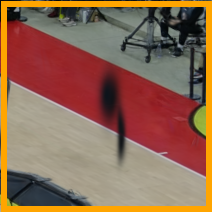} \\
        \includegraphics[width=0.98\linewidth]{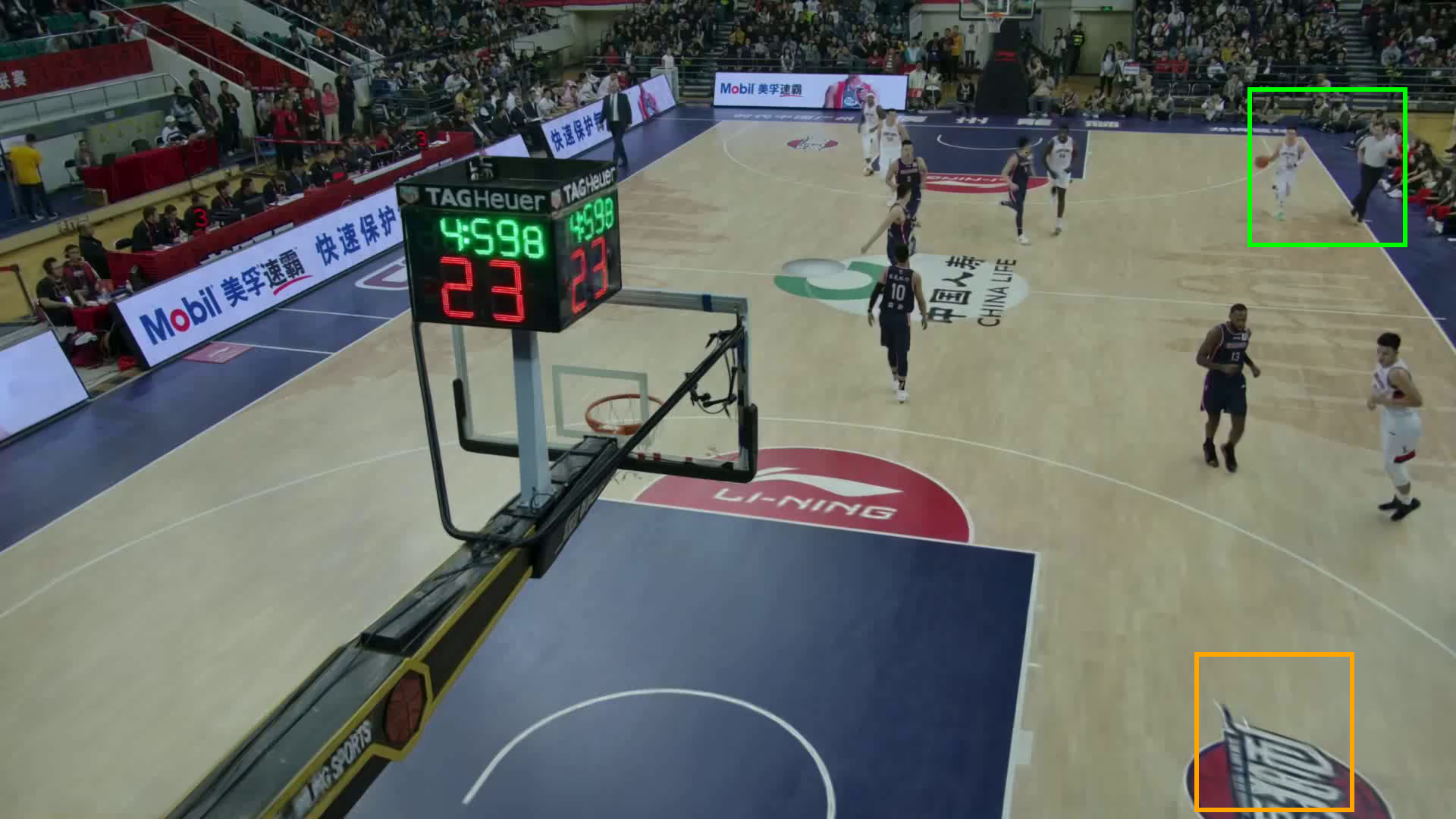}\\
        \includegraphics[width=0.475\linewidth]{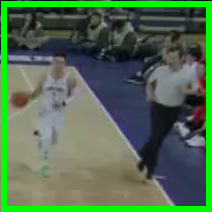}
        \includegraphics[width=0.475\linewidth]{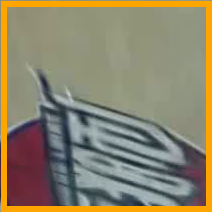}
        \caption*{(d) LocalDyGS \cite{Wu2025LocalDyGSMG}}
    \end{minipage}
    \caption{
    \textbf{Qualitative results on \textit{\vrulong} (1,400 frames; extreme motion amplitudes and long-sequence challenges) and \textit{VRU GZ} \cite{Wu2025LocalDyGSMG} (complex dynamic interactions).} Compared to 4DGaussian \cite{wu20244d} and LocalDyGS \cite{Wu2025LocalDyGSMG}, our method produces sharper renderings in both dynamic regions (e.g., athletes) and static areas (e.g., court floor), with more stable and temporally coherent reconstructions. Additional results are provided in the \textbf{supplementary video and materials}.
    }
    \label{fig:qua_1400}
    \vspace{-1.4em}
\end{figure*}

\section{Experiment}
\label{sec:experiment}

\subsection{Implementation}
\label{subsec:Implementation}
For our method, all MLPs consist of two layers with ReLU, followed by Sigmoid or normalization at the output. Both dynamic and static feature dimensions are set to 64. 


\begin{table}[]
\centering
\caption{\textbf{Quantitative comparison on \vrulong~, which contains 1400 frames.} Static methods are tested on frame 0. }
\vspace{-0.8em}
\scriptsize
\begin{tabular}{ccccc}
\toprule
Method      & PSNR $\uparrow$ & DSSIM$_1$ $\downarrow$ & DSSIM$_{2}$ $\downarrow$ & LPIPS $\downarrow$ \\ 

\midrule

\rowcolor{gray!20} 
\multicolumn{5}{l}{\textit{Static methods}}   \\
\midrule

2DGS \cite{huang20242d}        &  23.80                &  0.085                          &  0.042                          &  0.181                   \\
3DGS \cite{kerbl20233d}        &  24.13                &   0.087                         & 0.040                            & 0.159                    \\
ScaffoldGS \cite{lu2024scaffold}  & 23.53                 &  0.096                          & 0.048                           &  0.203                   \\ 

\midrule
\rowcolor{gray!20} 
\multicolumn{5}{l}{\textit{\streamframename~ methods}}   \\
\midrule

3DGStream \cite{sun20243dgstream}  &   21.94              &      0.105                     &  
0.053                        &        0.200            \\

iFVC \cite{tang2025compressing} & 22.35 & 0.101 & 0.050 & 0.192 \\ 

\midrule
\rowcolor{gray!20} 
\multicolumn{5}{l}{\textit{Clip methods}}   \\
\midrule

4DGaussian \cite{wu20244d}  &      22.01           &         0.103                  &    0.052                    &      0.198              \\

Swift4D \cite{wuswift4d}   & 23.01 & 0.094 & 0.048 & 0.180 \\

LocalDyGS \cite{Wu2025LocalDyGSMG}   & 23.11           & 0.093                     & 0.046                     & 0.178              \\

\midrule
\rowcolor{gray!20} 
\multicolumn{5}{l}{\textit{Clip-stream methods}}   \\
\midrule

\textbf{\ourname(Ours)}        & \textbf{24.54}  & \textbf{0.079}            & \textbf{0.036}                     & \textbf{0.146}     \\ \bottomrule
\end{tabular}
\label{table:qua_1400}
\vspace{-1.4em}
\end{table}

Our method use the Adam optimizer~\cite{kingma2014adam} with the learning rate schedule of 3DGS~\cite{kerbl20233d} within each clips. The learning rate scheduler is reinitialized at the start of each clip’s training, rather than inheriting the state from the \firstClip. Reinitializing the scheduler prevents the learning rate from becoming too small as training progresses, which would otherwise hinder effective optimization.









\subsection{Datasets}
\label{subsec:datasets}



\begin{table}[]
\centering
\scriptsize
\setlength{\tabcolsep}{12.5pt} 
\caption{\textbf{Quantitative comparison on VRU (\textit{GZ}) dataset \cite{wuswift4d}.} Static methods are evaluated on frame 0 only, serving as an upper-bound reference for dynamic reconstruction.}
\vspace{-1em}
\begin{tabular}{cccc}
\hline
\multicolumn{1}{c}{\textbf{Method}} & PSNR $\uparrow$ & SSIM $\uparrow$ & LPIPS $\downarrow$ \\ \hline
GOF \cite{yu2024gaussian}                                                  & 30.39           & 0.949           & 0.141              \\
2DGS \cite{huang20242d}                                                  & 30.78           & 0.949           & 0.187              \\
3DGS \cite{kerbl20233d}                                                  & 30.50           & 0.949           & 0.171              \\ \hline
4DGaussian \cite{wu20244d}                                                  & 28.32           & 0.930           & 0.186              \\
SpaceTimeGS \cite{li2024spacetime}                                          & 27.42           & 0.926           & 0.193              \\
LocalDyGS \cite{Wu2025LocalDyGSMG}                                             & 30.58           & 0.944           & 0.173              \\ \hline
\rowcolor{gray!20} 
\textbf{\ourname(Ours)}                                                  & \textbf{30.67}  & \textbf{0.946}  & \textbf{0.137}     \\ \hline
\end{tabular}
\label{table:qua_gz}
\vspace{-1.8em}
\end{table}

\begin{figure*}[h]
    \centering

    \begin{minipage}[b]{0.25\textwidth}
        \centering
        \includegraphics[width=0.98\linewidth]{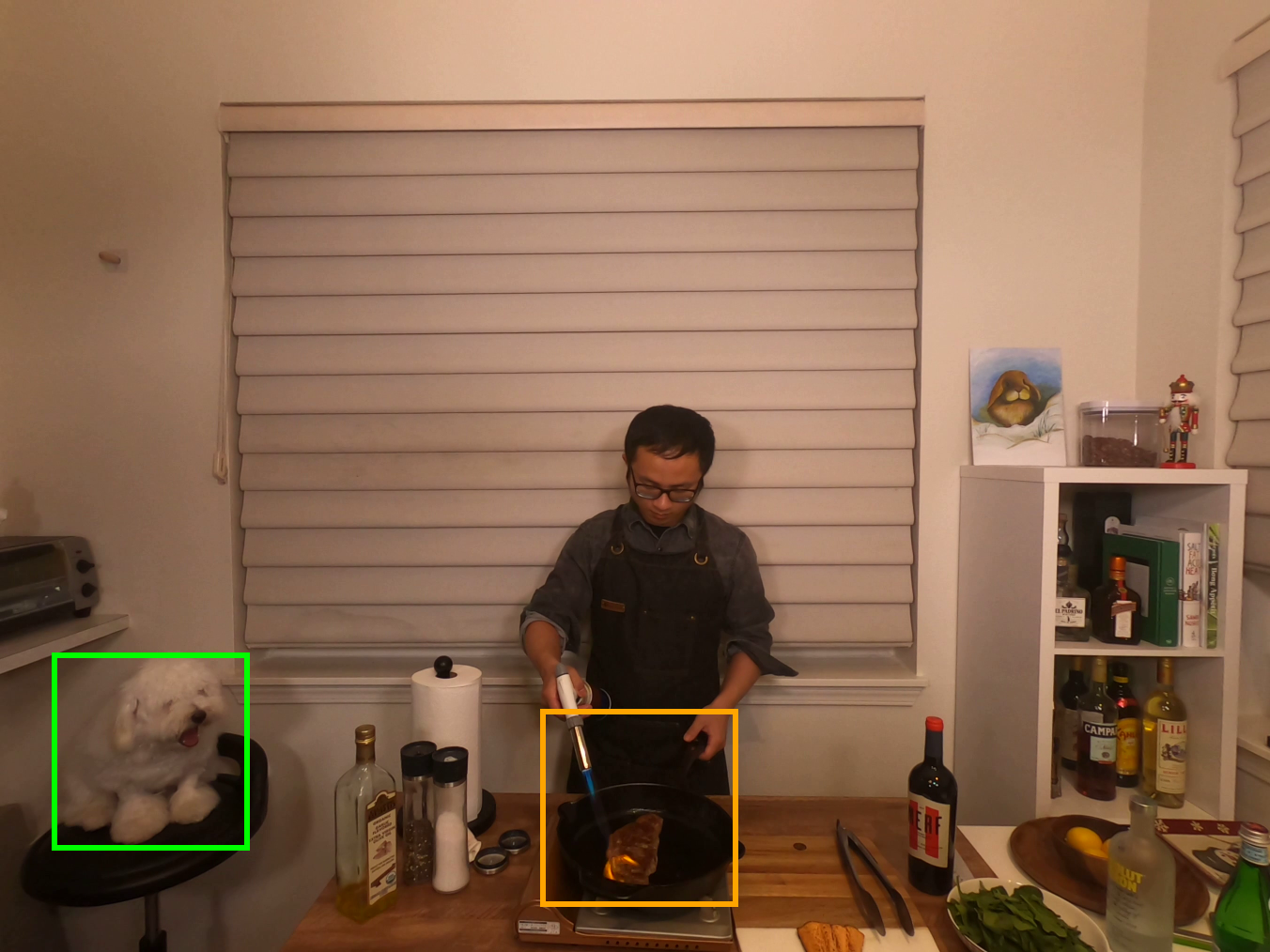} \\
        \includegraphics[width=0.475\linewidth]{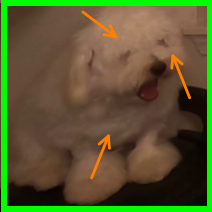}
        \includegraphics[width=0.475\linewidth]{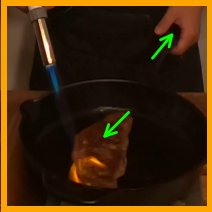}
        \caption*{(a) GT}
    \end{minipage}%
    \hfill
    \begin{minipage}[b]{0.25\textwidth}
        \centering
        \includegraphics[width=0.98\linewidth]{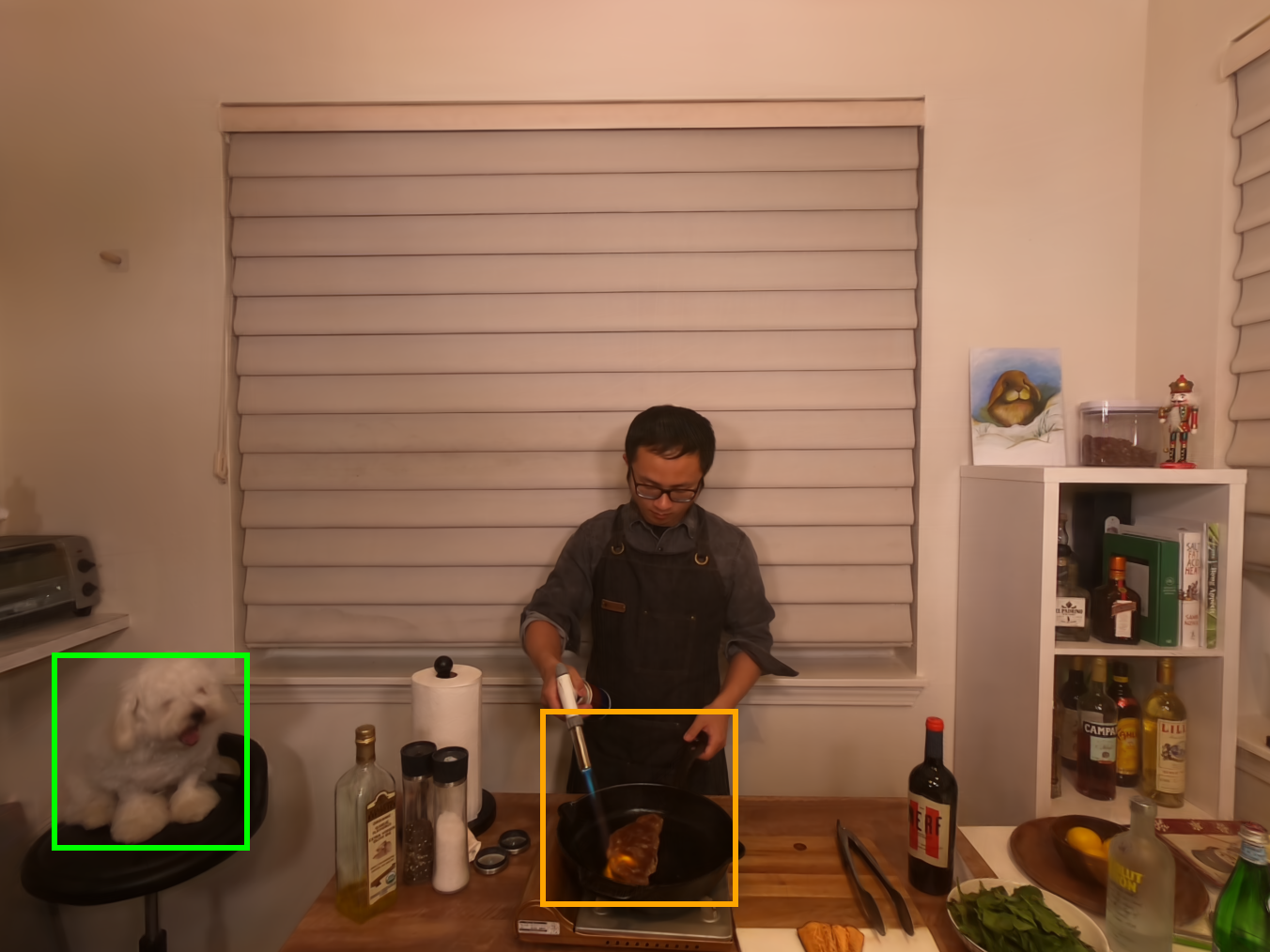} \\
        \includegraphics[width=0.475\linewidth]{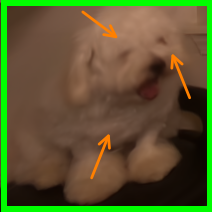}
        \includegraphics[width=0.475\linewidth]{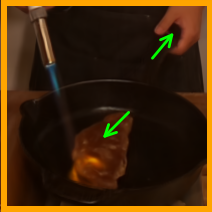}
        \caption*{(b) \textbf{Ours}}
    \end{minipage}%
    \hfill
    \begin{minipage}[b]{0.25\textwidth}
        \centering
        \includegraphics[width=0.98\linewidth]{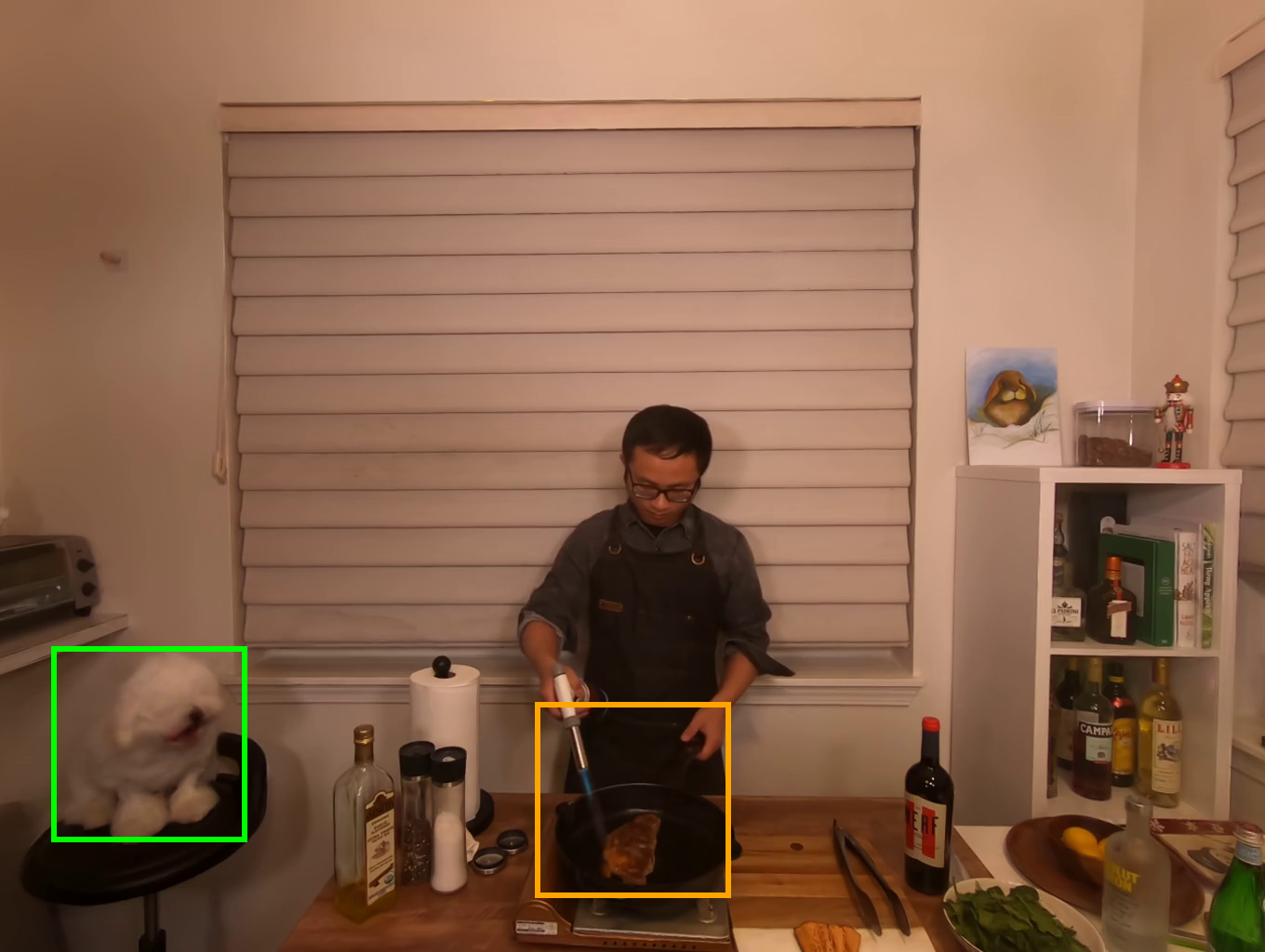} \\
        \includegraphics[width=0.475\linewidth]{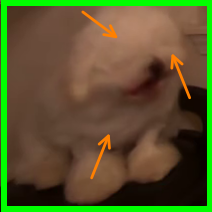}
        \includegraphics[width=0.475\linewidth]{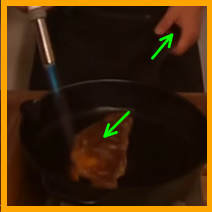}
        \caption*{(c) Grid4D \cite{xu2024grid4d}}
    \end{minipage}%
    \hfill
    \begin{minipage}[b]{0.25\textwidth}
        \centering
        \includegraphics[width=0.98\linewidth]{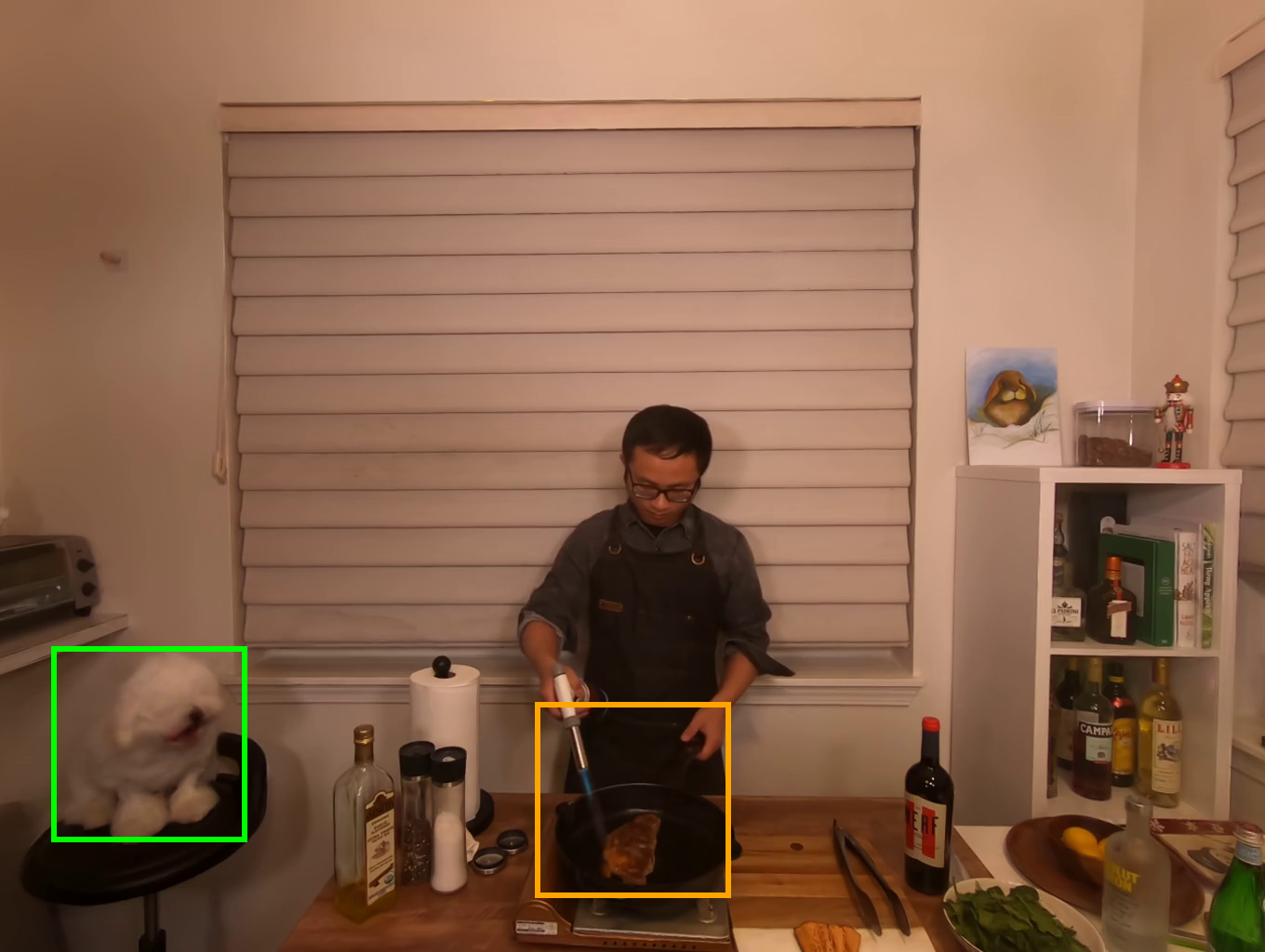}\\
        \includegraphics[width=0.475\linewidth]{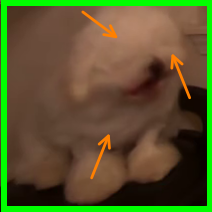}
        \includegraphics[width=0.475\linewidth]{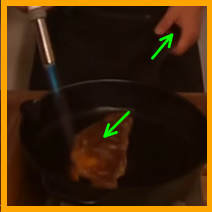}
        \caption*{(d) LocalDyGS \cite{Wu2025LocalDyGSMG}}
    \end{minipage}

    \vspace{-0.8em}
    \caption{\textbf{Qualitative results of \textit{flame salmon} from the N3DV dataset \cite{li2022neural}, a fine-scale motion sequence}. Our method outperforms SOTA approaches (Grid4D \cite{xu2024grid4d}, LocalDyGS \cite{Wu2025LocalDyGSMG}) by better preserving details in dynamic regions such as the dog’s face and flames. }
    \label{fig:qua_1200}
  	\vspace{-1.6em}
\end{figure*}
\textbf{\vrulong.} 
We primarily evaluate our method on the \vrulong~dataset \cite{Wu2025LocalDyGSMG, AVS}. 
 It consists of \textbf{1,400 frames} recorded at 4K resolution from 36 cameras uniformly arranged in a 360-degree circular setup, capturing a dynamic basketball game. This provides a challenging benchmark for reconstruction under high-speed motion. We use cameras 0, 10, 20, and 30 as test views, and the remaining 32 for training, with all images downsampled by a factor of 2.

\noindent
\textbf{N3DV~\cite{li2022neural}.} 
N3DV is a widely adopted benchmark, captured with a 21-camera multi-view system at a resolution of  2704×2028 and 30 FPS.
It is a standard dataset for multi-view dynamic reconstruction with fine-scale motion. The train/test settings follow the prior works~\cite{sun20243dgstream,wang2023mixed,li2022streaming}.


\noindent
\textbf{VRU~\cite{wuswift4d}.} 
The VRU dataset is captured with a 34-camera multi-view system, recording real-world basketball games at 1920×1080 resolution and 25 FPS. We follow the same experimental settings as Swift4D~\cite{wuswift4d}. 
Compared to N3DV, it features larger motion ranges and serves to evaluate robustness in real-world conditions.

\subsection{Comparisons}
\label{subsec:Comparisons}



 \begin{table}[]
 	\scriptsize
 	\setlength{\tabcolsep}{2pt}
 	\caption{\textbf{Quantitative comparisons are conducted on the Neural 3D Video Dataset \cite{li2021neural}, comprising five 300-frame scenes.}}
 	\vspace{-1em}
 	\resizebox{\linewidth}{!}{
 		\begin{tabular}{ccccccc}
 			\midrule
 			Method                                                                                      & PSNR $\uparrow$ & DSSIM$_{1}$ $\downarrow$ & DSSIM$_{2}$ $\downarrow$ & FPS $\uparrow$ & Time$\downarrow$ & Size$\downarrow$     \\ 
 			
 			\midrule
 			\rowcolor{gray!20} 
 			\multicolumn{7}{l}{\textit{\streamframename~methods}}   \\
 			\midrule
 			
 			StremRF \cite{li2022streaming}                           & 28.26           & -                        & -                        & 10.9           & -                & 5310MB               \\
 			3DGStream \cite{sun20243dgstream}                                & 31.67           & -                        & -                        & \cellcolor{red!25}215            & 1.0h             & 1230MB               \\
 			4DGC~\cite{Hu_2025_CVPR}                   & 31.58    & -                         & -                         & -  & - & -                \\
 			HiCoM~\cite{gao2024hicom}                &  31.17   & -                         & -                         & -  & - & -                \\

 			\midrule
 			\rowcolor{gray!20} 
 			\multicolumn{7}{l}{\textit{Clip methods}}   \\
 			\midrule

 			NeRFPlayer\cite{song2023nerfplayer}                                  & 30.69           & 0.034                    & -                        & 0.05           & 6.0h             & 5130MB               \\
 			HyperReel\cite{attal2023hyperreel}                                   & 31.10           & 0.036                    & -                        & 2              & -                & 360MB                \\
 			K-Planes\cite{fridovich2023k}                                        & 31.63           & -                        & 0.018                    & 0.3            & 5.0h             & 311MB                \\
 			HexPlane\cite{cao2023hexplane}                                       & 31.70           & -                        & \cellcolor{orange!25}0.014                    & 0.21           & 12.0h            & 240MB                \\
 			MixVoxels\cite{wang2022mixed}                                        & 31.73           & -                        & \cellcolor{yellow!25}0.015                    & 4.6            & -                & 500MB                \\
 			4DGaussian\cite{wu20244d}                                            & 31.02           & 0.030                    & -                        & 30             & \cellcolor{yellow!25}0.67h            & \cellcolor{red!25}90MB                 \\

 			RealTimeGS\cite{yang2023real}                                        & 32.01           & -                        & \cellcolor{orange!25}0.014                    & \cellcolor{yellow!25}114            & 9.0h             & \textgreater{}1000MB \\
 			SpaceTimeGS \cite{li2024spacetime}                                   & \cellcolor{yellow!25}32.05           & \cellcolor{orange!25}0.026                    & \cellcolor{orange!25}0.014                    & \cellcolor{orange!25}140            & \textgreater{}5h & 200MB                \\
 			LocalDyGS\cite{Wu2025LocalDyGSMG}                                    & \cellcolor{orange!25}32.28           & \cellcolor{yellow!25}0.028                    & \cellcolor{orange!25}0.014                    & 105            & \cellcolor{orange!25}0.58h            & \cellcolor{yellow!25}100MB                \\ 
 			
 			\midrule
 			\rowcolor{gray!20} 
 			\multicolumn{7}{l}{\textit{Clip-stream methods}}   \\
 			\midrule
 			\textbf{\ourname(Ours)} & \cellcolor{red!25}\textbf{32.53}  & \cellcolor{red!25}\textbf{0.024}           & \cellcolor{red!25}\textbf{0.012}           & 106   & \cellcolor{red!25}\textbf{0.5h}    & \cellcolor{orange!25}\textbf{98MB}        \\ \midrule
 		\end{tabular}
 	}
 	\vspace{-2em}
 	\label{table:qua_n3dv_short}
 \end{table}

\noindent
\textbf{Quantitative comparisons.} 
As shown in \cref{table:qua_1400}, \ourname~achieves the best performance across all objective metrics (PSNR, DSSIM$_1$, DSSIM$_2$, and LPIPS) on the \vrulong~dataset. 
\streamframename~methods struggle with large-scale motion due to their frame-wise dependency, while \nonframename~methods face optimization challenges when extended to long sequences. 
By combining the advantages of both paradigms, \ourname~achieves stable and high-fidelity reconstruction under large and complex motions.

\begin{table}[]
\centering
\scriptsize
\setlength{\tabcolsep}{12.5pt} 
\caption{\textbf{Quantitative comparison on \textit{\longNdv} (1200f).}}
\vspace{-1em}
\begin{tabular}{cccc}
\hline
Method     & PSNR $\uparrow$ & SSIM $\uparrow$ & LPIPS $\downarrow$ \\ \hline
4DGaussian \cite{wu20244d} & 28.89           & \textbf{0.952}           & 0.196              \\
4K4D \cite{xu20234k4d}      & 21.29           & 0.826           & 0.196              \\
ENeRF \cite{lin2022enerf}      & 23.48           & 0.894           & 0.259              \\
3DGS \cite{kerbl20233d}      & 28.61           & 0.949          & 0.210              \\
Dy3DGS \cite{luiten2023dynamic}     & 25.91           & 0.880           & 0.255              \\
LocalDyGS \cite{Wu2025LocalDyGSMG}  &  28.15               & 0.912                &  0.153                  \\ \hline
\rowcolor{gray!20} 
\textbf{\ourname(Ours)}       & \textbf{29.40}           & 0.917           & \textbf{0.144}              \\ \hline
\end{tabular}
\label{table:qua_1200}
\vspace{-1.6em}
\end{table}

We further validate the scalability of \ourname~on the \textit{flame salmon} sequence (1,200 frames) from N3DV, as reported in \cref{table:qua_1200}. 
To assess generalization, we also evaluate on the VRU \textit{GZ} sequence (250 frames) and other N3DV scenes (300 frames). 
Across all settings (\cref{table:qua_gz,table:qua_n3dv_short}), \ourname~consistently delivers state-of-the-art performance, demonstrating strong robustness and adaptability across diverse motion scales.

\noindent
\textbf{Qualitative comparisons.} 
As shown in Fig.~\ref{fig:qua_1400}, our method accurately reconstructs both the intricate motions of athletes and the fine details of the stationary floor. Furthermore, as shown in Fig. \ref{fig:qua_1200}, our method provides a more accurate representation compared to Grid4D and LocalDyGS, demonstrating its advantage in handling detailed motion.


\subsection{Ablation Studies}
\label{subsec: ablation}


%




\textbf{Feature Decomposition.}
 In \cref{fig:feat}, we render images by decoding only $f_s$ and only the dynamic features $f_d$ using the decoder. $f_s$ captures all background information within a clip, while $f_d$ learns residual information that controls the visibility of dynamic content. Thus, our inheritance strategy ensures the scene's temporal consistency while also enabling the modeling of dynamic content.




\noindent
\textbf{\sourceAnchorsCompensationModules.} During training, this module computes residual points as dynamic supplements based on anchors from \firstClip. It enhances dynamic fidelity by capturing fast motions (\cref{table:abl_optimizer}, w/o RAC) and removes redundant static anchors in \otherClip, mitigating flickering (\cref{fig:errorsMaps}(a)).


\begin{table}[t]
\scriptsize
\centering
\setlength{\tabcolsep}{10pt}
\setlength{\tabcolsep}{9pt} 
\caption{\textbf{Ablation on \decoderInheritModule~(DI) and \sourceAnchorsCompensationModules~(RAC) in \vrulong~. }}
\vspace{-1em}
\begin{tabular}{ccccc}
\hline
Method & PSNR $\uparrow$ & $\mathrm{DSSIM}_{1}$ $\downarrow$ & $\mathrm{DSSIM}_{2}$ $\downarrow$ & LPIPS $\downarrow$ \\ \hline
w/o DI                   & 24.34           & 0.081               & 0.038               & 0.152              \\
w/o RAC & 23.62 & 0.083 &	0.044 & 	0.160  \\ \hline
\textbf{ours}                        & \textbf{24.54}           & \textbf{0.079}               & \textbf{0.036}               & \textbf{0.146}              \\ \hline
\end{tabular}
\label{table:abl_optimizer}
\vspace{-1.4em}
\end{table}
\begin{figure}[t]
\centering
\setcounter{subfigure}{0}  
\subfloat[w/o optimizer inherited]{
		\includegraphics[scale=0.55]{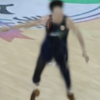}
		\includegraphics[scale=0.55]{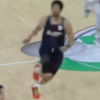}}
\subfloat[Full]{
		\includegraphics[scale=0.55]{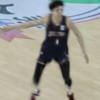}
		\includegraphics[scale=0.55]{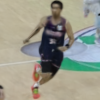}}
           \vspace{-0.8em}

\caption{\textbf{The Ablation study on decoder inheritance.} (a) Without inheriting the decoder, the rendered image exhibits noticeable blurriness. In contrast, employing decoder inheritance yields clear details, as visualized in (b). }
\label{fig:abl_optimizer}
\vspace{-1em}
\end{figure}

\noindent
\textbf{\clipIndepentSpatioTemporalFields.} 
Using a shared $STF$ across clips (Table~\ref{table:abl_smallGops}) causes later clips to overwrite previously learned dynamics. Training all clips independently leads to sparse anchors and degraded performance.

\noindent
\textbf{\decoderInheritModule.} We compare training \otherClipAbb~ with new decoders against inheriting the decoder from \firstClipAbb. Decoder inheritance preserves the consistency of decoded attributes across clips. As shown in \cref{fig:abl_optimizer}, this design significantly improves rendering quality in dynamic regions (e.g., basketball players) and achieves higher objective scores in \cref{table:abl_optimizer}.

\noindent
\textbf{\targetAnchorsInherit.} For each \otherClipAbb, we use anchors from both the \firstClipAbb~ and the current clip. Anchors and static features from \firstClipAbb~ are kept frozen to maintain consistent representation of static areas and prevent redundant optimization. When this constraint is removed, severe flickering artifacts appear across clips, as shown in \cref{fig:errorsMaps}(b). 

\begin{figure}[t]
	\centering

    \includegraphics[width=0.6\textwidth, trim=1cm 13cm 18cm 0.1cm]{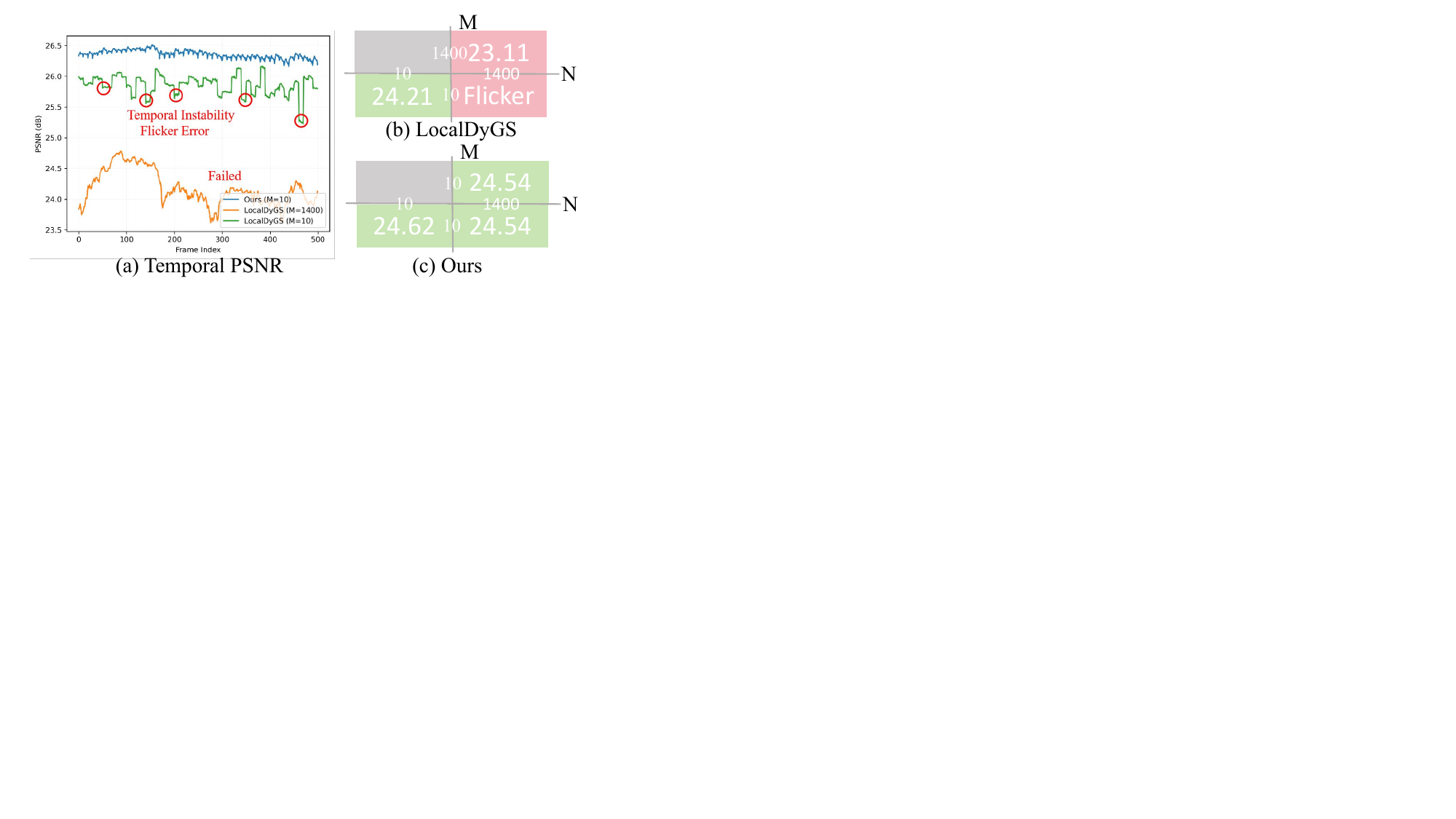}


	\caption{(a) Clip-based method LocalDyGS exhibits temporal instability with small $M$ and fail to train with larger $M$. In (b, c),
		our PSNR outperform LocalDyGS on both short and long sequences.}
	\vspace{-1.5em}
	\label{fig:MN}
\end{figure}

\noindent
\textbf{Ablation on $M$ and $N$.}
Given a total of $N$ input frames, the video is divided into segments of $M$ frames each. A critical limitation of prior methods is their inherent requirement for $M=N$, meaning the entire video sequence must be processed at once. Directly extending it to multiple clips by partitioning frames (i.e. $M<N$) results in noticeable temporal inconsistencies (\cref{fig:MN}, LocalDyGS 140 Clips video). Our method addresses this by enforcing consistency across clips, enabling flicker-free handling when $M<N$. 

\begin{table}[t]
\scriptsize
\centering
\setlength{\tabcolsep}{7.5pt} 

\caption{\textbf{Experiments on two clip training strategies.} Our method achieves superior objective quality. Experiments are carried out on the \vrulong~dataset.
}
\vspace{-1em}
\begin{tabular}{ccccc}
\hline
Method & PSNR $\uparrow$ & DSSIM $_1$ $\downarrow$ &DSSIM$_2$ $\downarrow$ & LPIPS $\downarrow$ \\ \hline
Independent Training                           & 21.85            & 0.142                    & 0.068                    & 0.316               \\
Shared STF & 23.11           & 0.093                   & 0.046                   & 0.178 \\ \hline
\textbf{ours}                             & \textbf{24.54}           & \textbf{0.079}                   & \textbf{0.036}                   & \textbf{0.146}              \\ \hline
\end{tabular}
\label{table:abl_smallGops}
\vspace{-1.4em}
\end{table}





\section{Conclusion}
\label{sec:conclusion}

We present \textbf{\ourname}, a unified training framework that bridges \streamframename~and \nonframename~dynamic Gaussian paradigms for high-quality modeling of long sequences and large-scale motion. Specifically, we uniformly divide long dynamic video sequences into multiple clips. Then, we propose a \dynamicIndepentTrainingStrategy~to effectively capture clip-specific dynamic features. To mitigate flickering artifacts across-clips, we further introduce a \staticInheritTrainingStrategy~that enforces temporal consistency across clips. Extensive experiments on \vrulong, VRU, and N3DV demonstrate that our framework achieves state-of-the-art performance in complex motion scenes and generalizes effectively to sequences of arbitrary length.

\noindent
\textbf{Limitation.} Our approach depends on camera poses estimated by COLMAP. In challenging cases with low image overlap or large textureless areas, COLMAP may yield inaccurate calibrations, which can affect reconstruction quality. Future work will focus on integrating more robust pose estimation to further improve generalization.

\section{Acknowledgments}
\label{sec:acknowledge}
This work is financially supported by Guangdong Provincial Key Laboratory of Ultra High Definition Immersive Media Technology(Grant No. 2024B1212010006), this work is also financially supported for Outstanding Talents Training Fund in Shenzhen, Shenzhen Science and Technology Program(Grant No. SYSPG20241211173440004 and RCJC20200714114435057), R24115SG MIGU-PKU META VISION TECHNOLOGY INNOVATION LAB.

{
    \small
    \bibliography{main}
}


\end{document}